\documentclass[a4paper,fleqn]{cas-dc}

\usepackage[numbers,sort&compress]{natbib}
\begin{document}
\let\WriteBookmarks\relax
\def\floatpagepagefraction{1}
\def\textpagefraction{.001}
\shorttitle{Failure Modes and Data-Efficient Remedies for Transformer-Based
Prediction of Viscoplastic Flows}
\shortauthors{M. Saberi et~al.}

\title [mode = title]{Failure Modes and Data-Efficient Remedies for Transformer-Based Prediction of Viscoplastic Flows}        

\author[1]{Maedeh Saberi}
% \ead{saberi.m@northeastern.edu}

\credit{Methodology, Investigation, Programming, Validation, Data Curation, Visualization, Writing (Original Draft)}

%\address[1]{, Street 129, 1043 NX Amsterdam, The Netherlands}
\affiliation[1]{organization={Department of Mechanical and Industrial Engineering},
                addressline={Northeastern University}, 
                city={Boston},
                postcode={02115}, 
                state={Massachusetts},
                country={USA}}

\author[2]{Hugo França}
\credit{Methodology, Programming, Validation, Data Curation, Writing (Original Draft)}

\author[2,3]{Maziyar Jalaal}
\credit{Methodology, Writing (Review and Editing)}

\author[1]{Safa Jamali}
\ead{s.jamali@northeastern.edu}
\credit{Supervision, Project administration, Funding acquisition, Methodology, Investigation, Writing (Review and Editing)}

\affiliation[2]{organization={Van der Waals-Zeeman Institute},
                addressline={Institute of Physics}, 
                addressline={University of Amsterdam}, 
                postcode={1098XH}, 
                % postcodesep={}, 
                city={Amsterdam},
                country={The Netherlands}}

\affiliation[3]{organization={Department of Applied Mathematics and Theoretical Physics},
                addressline={ University of Cambridge}, 
                city={Cambridge CB3 0WA},
                % postcode={825001}, 
                % state={Orissa}, 
                country={United Kingdom}}

\cortext[cor1]{Corresponding author}
% \cortext[cor2]{Principal corresponding author}
% \fntext[fn1]{This is the first author footnote, but is common to third
% \fntext[fn2]{Another author footnote, this is a very long footnote and
%   it should be a really long footnote. But this footnote is not yet
%   sufficiently long enough to make two lines of footnote text.}

% \nonumnote{This note has no numbers. In this work we demonstrate $a_b$
%   the formation Y\_1 of a new type of polariton on the interface
%   between a cuprous oxide slab and a polystyrene micro-sphere placed

\begin{abstract}
Fast and accurate prediction of viscoplastic flows remains a challenge in computational rheology due to the presence of yield surfaces, discontinuous gradients, and strongly nonlinear transitions between yielded and unyielded regions. 
Transformer-based neural operators such as RheOFormer have recently demonstrated strong potential for surrogate modeling of non-Newtonian flow problems.
However, their applicability to viscoplastic systems and their efficacy in recovering realistic flows remain poorly understood. 
In this work, we present the first systematic investigation of RheOFormer, applied to a simple flow of steady Bingham channel. By isolating the problem to a simple geometry and constitutive model, we examine in detail how dataset structure, numerical value range, and input representation govern the model’s ability to interpolate and extrapolate within the underlying parameter space. The data regimes in which RheOFormer remains accurate and those in which prediction errors emerge are identified, particularly for plug-dominated flows and for conditions outside the velocity scales or flow structures represented during training. Upon identification of these failure modes, targeted enhancements to the input representation and loss formulation are introduced to improve the prediction of low-gradient plug regions. Logarithmic velocity representations and gradient-aware loss formulations substantially improve accuracy in several challenging regimes where the baseline formulation becomes inadequate. In addition, extrapolation errors are reduced by augmenting the training set with only a small number of strategically selected cases spanning the missing physical regimes. 
Finally, a single partially developed snapshot is found to be sufficient for an accurate steady state prediction, and the model outperforms simply copying its input throughout the development of the flow.
These results highlight the importance of appropriate field representation and parameter-space coverage for data-efficient neural-operator prediction of viscoplastic flows.

\end{abstract}

% \includegraphics{figs/cas-grabs.pdf}

% \begin{highlights}
% \item Research highlights item 1
% \item Research highlights item 2
% \item Research highlights item 3
% \end{highlights}

\begin{keywords}
Surrogate Modeling \sep Viscoplastic Flows \sep Neural Operators \sep Transformer-Based Modeling
\end{keywords}

\maketitle

\section{Introduction}\label{sec1}

Yield-stress fluids exhibit a dual mechanical response: they deform as fluids when the local stress exceeds a critical value but remain nominally unyielded below that threshold. This behavior is encountered in materials ranging from pastes, cementitious suspensions, drilling fluids, and waxy crude oils to foods, biological materials, and printable inks \cite{balmforth2001geophysical, blake1990viscoplastic,steffe1996rheological, bonn2017yield}. 
Their flow therefore differs fundamentally from that of Newtonian fluids. 
The domain is partitioned into yielded regions with finite deformation rates and unyielded regions that move approximately as rigid plugs, with the boundaries between them determined as part of the solution \cite{dean2007numerical, coussot2014yield}. Even in geometrically simple configurations, changes in forcing or material properties can alter the location and extent of these regions, producing solution families that involve both continuous changes in velocity scale and qualitative changes in flow structure \cite{balmforth2025implications, papanastasiou1987flows,huilgol2019start}.

The Bingham model provides the canonical continuum description of this behavior by combining a yield stress with a post-yield plastic viscosity \cite{bingham1922fluidity, frigaard2017bingham}. 
Despite its apparent simplicity, numerical simulation of Bingham flow remains nontrivial because the ideal constitutive law is nonsmooth at yielding and the yield surfaces are not known a priori \cite{roberts2023predicting, balmforth2025implications}. 
Computational treatments commonly employ regularized constitutive equations, variational inequalities, augmented-Lagrangian methods, or related approaches that avoid explicit tracking of the yielded–unyielded interface \cite{saramito2017progress, mitsoulis2017numerical}.
Regularization makes the problem computationally tractable but introduces additional numerical scales and replaces perfectly rigid regions with zones of very small deformation rate. Comparisons against augmented-Lagrangian solutions indicate that this substitution can shift the apparent yield-surface location in some configurations \cite{frigaard2005usage}. Consequently, accuracy cannot be assessed through global velocity field alone: the predicted plug velocity, plug width, yield-surface position, wall shear, and flow rate can remain sensitive to spatial resolution, regularization, and solver tolerances. Pressure-driven channel flow is particularly useful in this context because it retains the essential yielded–unyielded structure while providing a controlled setting in which numerical and data-driven predictions can be systematically evaluated \cite{roberts2023predicting}.

High-fidelity simulations of viscoplastic flows must generally be repeated as the Reynolds number ($Re$), Bingham number ($Bi$), geometry, boundary conditions, or constitutive parameters change\cite{balmforth2014yielding,jalaal2021spreading,xie2016lattice}. 
This repeated-solution requirement can become expensive in parameter studies, optimization, uncertainty quantification, and real-time process control. Scientific machine-learning methods offer a possible route to amortizing this cost by learning surrogate mappings from material and flow conditions to the corresponding solution fields \cite{brunton2019data}. 
Physics-informed neural networks have been extended to non-Newtonian and elastoviscoplastic systems by embedding the conservation and constitutive equations into the training objective \cite{brunton2020machine,mahmoudabadbozchelou2021rhinns,lennon2023scientific,mangal2025data}. 
These methods have demonstrated considerable promise for forward prediction, inverse parameter estimation, and recovery of rheological behavior from limited observations \cite{lardy2025inferring, mahmoudabadbozchelou2021rhinns}.
Variants such as non-Newtonian PINNs \cite{mahmoudabadbozchelou2022nn}, fractional PINNs \cite{dabiri2025detailed} and non-local PINNs \cite{zolfaghari2026non} have extended these capabilities to flow prediction and parameter discovery for a range of rheological systems. 
ViscoelasticNet goes further still, inferring stress fields from velocity data alone while selecting among Oldroyd-B, Giesekus, and Phan--Thien--Tanner responses
\cite{thakur2024viscoelasticnet}.
Viscoplastic flows have themselves received attention within this framework. Papanastasiou-regularized and augmented-Lagrangian PINN formulations have been used to predict and reconstruct steady Bingham flow fields \cite{gao2024deep, zhang2023physics}, and a convolutional network has been trained to reconstruct viscoplastic velocity fields and locate yield surfaces in superhydrophobic channels \cite{joulaei2025deep}. 
Lardy et al.\ further inferred viscoplastic constitutive laws, including the Herschel--Bulkley and Papanastasiou models, directly from velocity fields by jointly optimizing a neural representation of the flow and the constitutive model against the governing equations \cite{lardy2025inferring}.
These studies demonstrate that learned models can recover plug structure in specific viscoplastic configurations. However, these approaches remain largely instance-specific, with separate models trained for fixed configurations rather than a single mapping that transfers across the material and flow parameters defining the problem. Despite these successes, conventional neural network models face significant challenges when applied to strongly non-Newtonian flows. Sharp transitions, stiff nonlinearities, and long-horizon temporal dependencies degrade their accuracy, while their reliance on pointwise loss formulations makes it difficult to capture the global operator structure of the underlying PDEs. 

Motivated by these limitations, neural operators represent a more recent paradigm aimed at overcoming these challenges by learning the solution operator of a partial differential equation rather than predicting individual states for fixed problem configurations. Architectures such as DeepONet \cite{lu2021learning} and the Fourier Neural Operator (FNO) \cite{li2020fourier} have demonstrated strong performance across a range of smooth systems, including Navier–Stokes and diffusion \cite{kovachki2023neural}. Their accuracy is not uniform across problem types, however. Systematic benchmarking has shown that operator-learning performance can degrade on fields containing sharp or localized structure, while FNO-type architectures in particular attenuate high-wavenumber content by construction through spectral truncation \cite{lu2022comprehensive}. This is one manifestation of spectral bias, the tendency of neural networks to fit smooth global structure before localized features \cite{rahaman2019spectral}. The consequence is that thin layers and abrupt transitions are systematically oversmoothed, which is directly relevant to flows containing yield surfaces. Within rheology, Mangal et al. demonstrated that DeepONet- and FNO-based models can learn families of thixotropic elastoviscoplastic and Giesekus constitutive responses under varied flow protocols \cite{mangal2025learning}. 
Transformer-based neural operators further extend this capability through attention mechanisms that enable the model to capture global spatial correlations and long-range dependencies in the solution field \cite{li2022transformer,hao2023gnot}. Building on these advances, RheOFormer was recently introduced as a transformer-based neural operator tailored for rheological systems, with initial studies focusing primarily on viscoelastic flows \cite{saberi2025rheoformer}. However, its applicability and limitations for viscoplastic fluids remain largely unexplored. Yield surfaces, discontinuous gradients, and extended plug regions introduce numerical features distinct from those encountered in the viscoelastic systems previously considered with RheOFormer, raising open questions about its ability to accurately represent and generalize across such regimes.

A second consideration is the amount and distribution of training data required for reliable operator learning. Purely data-driven operators generally perform best within the distribution represented during training, while accuracy can degrade substantially under out-of-distribution prediction. Zhu et al. \cite{zhu2023reliable} quantified this for DeepONets by relating extrapolation error to a distance between training and test function spaces, and proposed fine-tuning and multifidelity strategies that restore accuracy when either the governing equations or sparse additional observations are available \cite{zolfaghari2026multi,zolfaghari2026hierarchical}. This framing raises a question specific to viscoplastic flows. Because $Re$ and $Bi$ affect the solution in qualitatively different ways, one rescaling the velocity magnitude and the other reorganizing the flow into yielded and unyielded regions, moving outside the training range along the two axes need not produce the same kind of failure or require the same additional training information.

Here, we use pressure-driven Bingham channel flow as a controlled benchmark to identify the conditions under which RheOFormer succeeds or fails when predicting steady viscoplastic flow from limited early-time information.
This flow is chosen because its exact steady solution is known and its yield surfaces are flat and fixed.
The objective is not to introduce a new neural-operator architecture, but to separate the effects of training-set size, numerical representation, and parameter-space coverage. The study is carried out in different stages. First, the dependence of prediction accuracy of the baseline formulation on the number and distribution of training cases is quantified. We then examine a systematic failure of the baseline formulation in plug-dominated flows, in which the plateau velocity is underpredicted, and test whether velocity-gradient inputs, logarithmically scaled representations, low-shear-weighted objectives, and Sobolev-type losses improve the reconstruction of the plug velocity and yielded–unyielded transition. Next, we distinguish shifts in velocity magnitude across the $Re$ range from changes in profile structure across the $Bi$ range. Finally, we determine whether a small number of strategically selected anchor cases spanning the Newtonian-like and plug-dominated limits can convert out-of-distribution failures into accurate interpolation over an expanded parameter domain. We also examine how the number and timing of the input snapshots affect the prediction, and compare the model with the trivial predictor that copies its last input. This analysis provides a physically interpretable account of how field representation and regime coverage govern the accuracy and data-efficiency of neural-operator prediction for viscoplastic flow.

In the following sections, Section~\ref{sec:methods} describes the dataset generation, model architecture, and training objectives, with the governing equations and numerical details given in Appendix~A.
Section~\ref{sec:results} then presents a detailed analysis of model performance, including interpolation, extrapolation behavior, failure modes, and the corresponding methodological improvements.

\section{Methods} \label{sec:methods}

\subsection{Training dataset}
\label{sec:data}
To generate the training dataset, two-dimensional pressure-driven Bingham flows were simulated using the Basilisk flow solver. As described in detail in Appendix~\ref{section:app:numerical_details_validation}, our simulations are defined by three parameters only: $Re$, $Bi$ and $\bar{K}$. To systematically sample different rheological regimes and velocity scales, simulations are performed for nine Reynolds numbers,
\begin{equation}
    Re=
    \{0.01,0.025,0.05,0.075,0.1,0.25,0.5,0.75,1\}.
\end{equation}
For each Reynolds number, 21 values of $Bi$ are considered, uniformly distributed between the Newtonian limit $Bi=0$ and $Bi=8Re$. The nondimensional pressure gradient is kept fixed in all simulations as $\bar{K}=10$. This parameter range spans flows from Newtonian behavior to strongly viscoplastic regimes with substantial unyielded plug regions. See Appendix~\ref{section:app:numerical_details_validation} for additional details of the equations being solved and the numerical scheme used to obtain their solutions.
Equation~\eqref{eq:analytical_solution} shows that, for fixed $\bar{K}$, the steady profile can be written as $u_x(y) = Re\, g(y;\, Bi/Re)$, so that $Re$ only rescales the velocity magnitude, while the profile shape is set by $Bi/Re$ alone.
In the present dataset, varying $Re$ at fixed $Bi/Re$ therefore probes the model's response to changes in velocity scale, whereas varying $Bi/Re$ probes its response to changes in flow structure.
We deliberately use the velocity fields in unnormalized form, since extrapolation in velocity scale is one of the failure modes under study.

% =====================================================
% END OF NEW VERSION WRITTEN BY HUGO
% =====================================================

The dataset is used in several different train–test configurations. In particular, we consider multiple training scenarios with varying dataset sizes and parameter-space coverage, allowing us to separately assess interpolation accuracy, extrapolation to unseen $Bi$ or $Re$, and sensitivity to the amount of temporal input information. The specifics of each split are reported alongside the corresponding results.
All representative profiles shown in the figures are held-out cases that are not used for training.

In the remaining sections of this paper, all variables are dimensionless. For convenience, we omit the upper bars introduced in the nondimensionalization of Eq.~\eqref{eq:rescalings}. 

\begin{figure*}[t]
\centering
\includegraphics[width=\textwidth]{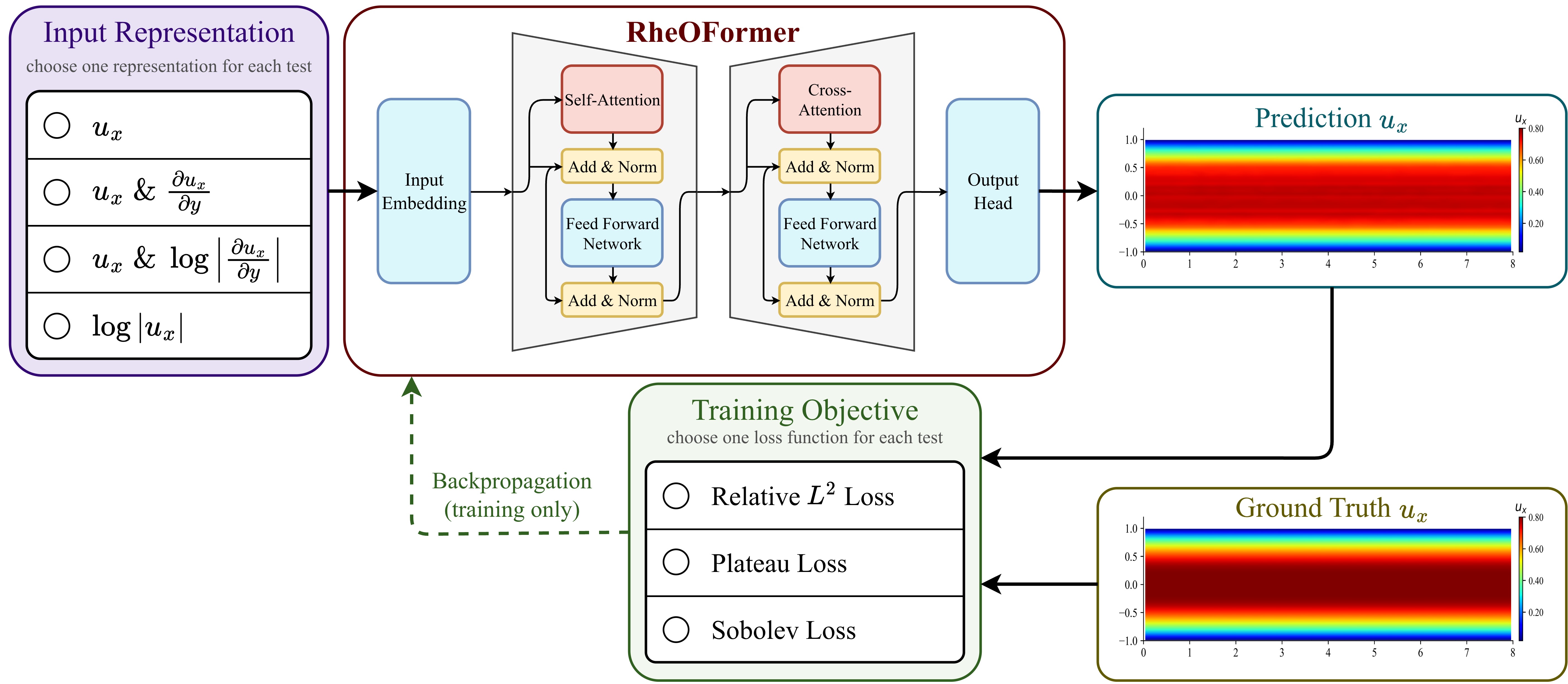}
\caption{Schematic of the RheOFormer framework as applied to steady Bingham channel flow. The model receives a short sequence of early-time velocity snapshots in one of four representations (left). 
The inputs are projected by a feed-forward embedding, encoded through self-attention layers with residual connections and layer normalization, and decoded by cross-attention. 
The output head returns the predicted steady state field $u_x(x,y)$. During training (dashed line), the prediction is compared against the simulated steady state under one of three objectives. Each experiment reported in Section~\ref{sec:results} uses one representation and one objective; all other architectural and optimization settings are held fixed. The fields shown are for ${Re}=0.25$, ${Bi}=0.5$.
}
\label{fig:architecture}
\end{figure*}

\subsection{RheOFormer Architecture}

RheOFormer is a transformer-based neural operator specifically developed to approximate nonlinear solution operators arising in rheological partial differential equations (PDEs) \cite{saberi2025rheoformer}. Its design is intended to learn mappings between input and output fields across varying flow conditions rather than fitting individual trajectories independently. Originally introduced as a general framework for modeling complex fluid dynamics, RheOFormer integrates attention-based encoding \cite{vaswani2017attention} with coordinate-driven decoding \cite{rahimi2007random} to capture both memory effects and long-range spatial correlations that commonly appear in viscoplastic and non-Newtonian flows. Figure~\ref{fig:architecture} summarizes the framework, including the input representations and training objectives compared in Section~\ref{sec:results}.

In our application, the model receives as input a limited number of early time spatial fields corresponding to the initial snapshots of the velocity component $u_x(x,y,t)$, along with their associated spatial coordinates.  These snapshots are first projected into a latent representation through a feed-forward embedding network that preserves both local and global spatial information. This latent representation is subsequently processed by multiple layers of self-attention, enabling the model to identify the dominant spatial couplings that govern the evolution of the flow. Residual connections and layer normalization are incorporated throughout these layers, which ensures numerical stability and facilitates efficient optimization as the network depth increases.

Within each attention layer, every spatial point interacts with all others, allowing the model to learn highly nonlocal spatial dependencies that are essential for representing Bingham-type behavior. 
Although the Bingham constitutive law is local, the resulting velocity field and yield-surface location are determined through the domain-wide momentum balance, so the plug extent at one location depends on conditions throughout the domain.
Mathematically, given a set of embedded input tokens $X \in \mathbb{R}^{N \times n}$, self-attention constructs queries, keys, and values via linear projections $Q = XW_Q$, $K = XW_K$, and $V = XW_V$. The attention output is computed as

\begin{equation}
\mathrm{Attention}(Q,K,V) = \frac{Q\left(K^\top V\right)}{n},
\end{equation}
which defines a global operator where each point aggregates information from all other points weighted by their pairwise similarity. Multi-head attention extends this mechanism by learning several such operators in parallel, thereby capturing multiple interaction scales and directional couplings that arise in viscoplastic flows.

To produce predictions on arbitrary spatial grids, RheOFormer employs a coordinate-based decoder that separates the output resolution from the resolution of the input snapshots. Spatial coordinates are embedded using random Fourier features \cite{tancik2020fourier}, improving the network’s ability to represent high frequency structures such as sharp velocity gradients and yield surfaces. A cross-attention mechanism then links the encoded latent representation to these target coordinates, enabling the model to evaluate the learned operator at any desired spatial location, decoupling the output resolution from the input resolution. The final feed-forward module maps the decoded latent features back to the physical field quantities.

In the present study, we adapt this architecture to the task of predicting the steady state velocity field in the Bingham channel-flow dataset. By learning a mapping from a small set of early-time snapshots to the final steady state distribution, RheOFormer effectively reconstructs the long-time rheological behavior without explicitly simulating the full transient evolution. Formally, the model approximates $\mathcal{G}: \{u_x(x,y,t_i)\}_{i=1}^{k} \mapsto u_x^{\infty}(x,y)$, and all errors reported below are computed on this single output field. Once trained, the steady state is obtained in a single forward pass without explicit integration of the remaining transient dynamics. This operator learning approach significantly reduces computational cost while retaining the ability to capture the nonlinear convergence dynamics characteristic of viscoplastic flows.

\subsection{Training Setup}

Four representations of the velocity data are compared. The baseline uses the raw streamwise velocity $u_x$ as a single input channel. The second representation augments $u_x$ with the transverse gradient $\partial u_x/\partial y$, providing explicit shear-rate information for distinguishing yielded and unyielded regions. The third replaces that second channel with $\log|\partial u_x/\partial y|$, which expands the numerical range of small gradient magnitudes clustered near zero within the plug. The fourth applies an analogous transformation to the velocity field, mapping $u_x \mapsto \log|u_x|$ on both the input and the target, so that the network is trained and evaluated in logarithmic space and predictions are mapped back by exponentiation before errors are computed. 
To avoid the singularity of the logarithm at the no-slip walls, in the quiescent initial condition, and inside the plug, a small constant $\delta = 10^{-8}$ is added before the transformation, i.e. $\log(|u_x| + \delta)$ and $\log(|\partial u_x/\partial y| + \delta)$; this sets a floor on the transformed values at these locations.
In every case the target quantity is the steady state field; only its representation during training differs. Reported errors are always computed on the physical velocity field, so values are directly comparable across representations.

In all baseline experiments, the model receives the first five temporal snapshots of the velocity field as input, spaced by $\Delta t = 0.03$ (input window $t=0$--$0.12$), while the target is the steady state. This choice reflects a balance between providing sufficient early-time information and limiting input dimensionality, enabling the model to infer the eventual steady state structure of the flow. To systematically evaluate the influence of temporal information, we also conduct experiments in which the number of input frames is varied from one to five and the input window is shifted in time (Section~\ref{sec:temporal}). Across all cases, the model receives no explicit values of $Re$ or $Bi$; instead, it is required to infer the underlying flow regime directly from the spatial patterns present in the input velocity fields alone. This constraint tests whether the operator recovers regime information from field structure rather than interpolating a parameter label, and it matches the practical setting in which early-time measurements are available but material parameters are not independently known.

Training is supervised using a relative $L_2$ loss computed between the predicted and ground truth steady state velocity fields. 
For a predicted steady state velocity field $u_{\mathrm{pred}}$ and the corresponding ground truth field $u_{\mathrm{gt}}$, the relative loss for a batch of size B is defined as
\begin{equation}
\mathcal{L}_{\mathrm{rel}}
=
\frac{1}{B}
\sum_{b=1}^{B}
\left(
\frac{
\left\|\,u_{\mathrm{pred}}^{(b)} - u_{\mathrm{gt}}^{(b)}\,\right\|_{2}
}{
\left\|\,u_{\mathrm{gt}}^{(b)}\,\right\|_{2} + \varepsilon
}
\right),
\end{equation}
where the $L_2$ norms are computed jointly over all spatial nodes of the steady state field, and $\varepsilon$ is a small regularization constant added to avoid division by zero. 
This normalization balances contributions across samples that differ significantly in magnitude due to variation in $Re$–$Bi$ conditions, thereby preventing the optimization from being dominated by high magnitude velocity values.

To further investigate whether incorporating derivative information improves predictive accuracy, we additionally employ a Sobolev training loss \cite{czarnecki2017sobolev} that combines relative velocity and relative gradient errors within a unified objective.
Let $\nabla u_{\mathrm{gt}}$ be the ground truth gradient directly obtained from the dataset. The gradient of the prediction, $\nabla u_{\mathrm{pred}}$, is computed numerically during training. 
Both gradients are computed using the same numerical differentiation scheme.
For a batch of size $B$, the Sobolev loss is defined as

\begin{equation}
\begin{aligned}
\mathcal{L}_{\mathrm{Sobolev}}
= \frac{1}{B}\sum_{b=1}^{B} \Bigg(
& W_1
\frac{
\left\|u_{\mathrm{pred}}^{(b)} - u_{\mathrm{gt}}^{(b)}\right\|_2
}{
\left\|u_{\mathrm{gt}}^{(b)}\right\|_2 + \varepsilon
}
\\
&+ W_2
\frac{
\left\|\nabla u_{\mathrm{pred}}^{(b)} - \nabla u_{\mathrm{gt}}^{(b)}\right\|_2
}{
\left\|\nabla u_{\mathrm{gt}}^{(b)}\right\|_2 + \varepsilon
}
\Bigg),
\end{aligned}
\end{equation}
where all norms are computed jointly over spatial nodes, and $\varepsilon$ is a small regularization constant preventing division by zero. The weights $W_1$ and $W_2$ control the relative importance of matching the velocity field and its spatial gradients, and are set to $W_1=10$ and $W_2=0.7$ in all experiments. 
In this formulation, the model is penalized not only for discrepancies in the predicted velocity field but also for mismatches in its spatial gradients, thereby encouraging more accurate reconstruction of shear layers and yield surfaces.

Building on the insight that regions with very small velocity gradients are particularly challenging for neural operators to reconstruct, we further incorporate a plateau-focused reweighting strategy. This approach directly targets the unyielded plug regions that are central to viscoplastic flow behavior. 
The loss selectively amplifies the contribution of grid points where the ground truth gradient indicates low-shear or nearly unyielded behavior. 
To identify such regions, we compute a mask $M(x,y)\in\{0,1\}$ directly from the ground truth velocity gradients. 
For each sample, a threshold $\tau_{\mathrm{th}}$ is computed from the distribution of ground truth gradient magnitudes, and points satisfying $|\nabla u_{\mathrm{gt}}|\leq\tau_{\mathrm{th}}$ are classified as plateau-region points.
Formally, letting $\nabla u_{\mathrm{gt}}$ denote the ground truth gradient and $\tau_{\mathrm{th}}$ the threshold, the mask is defined as

% \begin{equation}
% M(x,y)
% =
% \mathbf{1}\!\left(
% \,|\nabla u_{\mathrm{gt}}(x,y)|\; \le\; \tau_{\mathrm{th}}
% \right).
% \end{equation}

\begin{equation}
\label{eq:mask}
M(x,y)
=
H\!\left(
\tau_{\mathrm{th}} -
\left|\nabla u_{\mathrm{gt}}(x,y)\right|
\right),
\end{equation}
where $H(\cdot)$ denotes the Heaviside step function with $H(0)=1$.

The plateau loss then increases the weight of squared error contributions inside these masked regions. For a batch of size $B$, this loss takes the form
\begin{equation}
\small
\mathcal{L}_{\mathrm{plateau}}
=
\frac{1}{BN}
\sum_{b=1}^{B}\sum_{n=1}^{N}
\left[1+\lambda M^{(b,n)}\right]
\left(u_{\mathrm{pred}}^{(b,n)}-u_{\mathrm{gt}}^{(b,n)}\right)^2
\end{equation}
where $\lambda>0$ controls the additional penalty applied to plateau points.  
This formulation is intended to increase the contribution of low-shear regions during optimization and thereby encourage reconstruction of the extent and uniformity of the unyielded zones. 
In all experiments, $\tau_{\mathrm{th}}$ is set in each sample to $10\%$ of the range of the velocity gradient, $\tau_{\mathrm{th}} = 0.1 \operatorname{range}(\nabla u_{\mathrm{gt}})$, and $\lambda=1$, so that plateau points receive twice the weight of the remaining points.

Model optimization is performed using the AdamW optimizer (weight decay $10^{-4}$) with a one-cycle learning-rate schedule: the learning rate rises from $4\times10^{-6}$ to a peak of $10^{-4}$ over the first $10\%$ of training and then decays to approximately $4\times10^{-9}$, for 1000 epochs. 
All experiments use the same architecture, with 8 self-attention layers, one attention head, and a latent width of 256.
Batches contain up to 16 samples, so training sets smaller than this are processed full-batch. Each configuration is trained with three random seeds; the figures show one of these runs, and the averaged errors quoted in the text are means over the three seeds. 
Experiments are conducted under multiple dataset configurations to systematically assess interpolation, extrapolation across the $Re$--$Bi$ parameter space, and sensitivity to the amount of temporal input information.

\section{Results} \label{sec:results}

The ultimate goal of this work is to evaluate the ability of RheOFormer to infer the steady state behavior of viscoplastic channel flows from limited temporal information. Hence we seek to first demonstrate that the model successfully learns the underlying steady state physics across a range of flow regimes; and second, to probe the limits of this capability through controlled variations in the dataset and inputs. To this end, results are presented in a sequence of targeted validations, each designed to isolate a specific aspect of model performance, as detailed in the subsections below.

\subsection{Steady State Prediction Across $Re$--$Bi$ Parameter Space}

\begin{figure*}
\centering
\includegraphics[width=\textwidth]{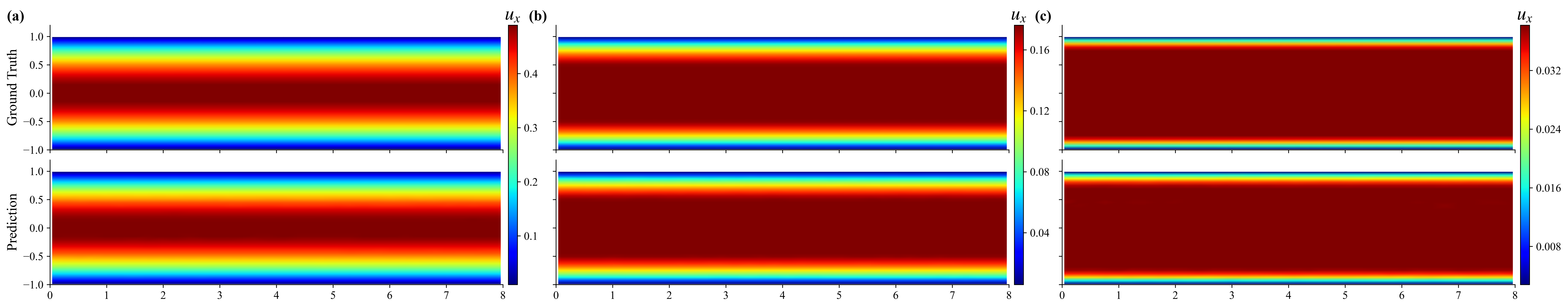}
\caption{Comparison of ground truth data (\textbf{top row}) vs. RheOFormer predictions (\textbf{bottom row}) for steady state streamwise velocity fields at $Re=0.1$ and three representative $Bi$. From left to right: 
(a) $Bi=0.00$ (Newtonian limit), 
(b) $Bi=0.40$ (moderate viscoplastic behavior), and (c) $Bi=0.72$ (strong plug region). Color scales differ between cases for clarity.
}
\label{fig:steady_state_examples}
\end{figure*}

Before assessing the failure modes of the model with respect to data availability, RheOFormer's ability to predict steady state velocity fields across the full range of $Re$  and $Bi$ in the dataset is studied. 
In this baseline study, the model is trained using the full dataset, and its performance is evaluated on previously unseen flow conditions that lie within the parameter range spanned by the training data.
Figure~\ref{fig:steady_state_examples} shows three representative cases at $Re=0.1$ with increasing $Bi$, illustrating the transition from Newtonian-like flow ($Bi=0$) to strongly viscoplastic behavior with a well defined plug region ($Bi=0.72$). In all cases, the model receives only the first five snapshots of the time evolution ($t = 0$--$0.12$) and is tasked with predicting the final steady state distribution.

For the Newtonian limit ($Bi=0$), the predicted velocity field matches the ground truth almost exactly, capturing both the shape and magnitude of the parabolic profile. This agreement indicates that RheOFormer successfully learns the governing structure of shear dominated flows and their convergence toward steady state. As the $Bi$ increases ($Bi=0.40$), the velocity profile begins to flatten due to the presence of an emerging unyielded core. RheOFormer captures this transition with high fidelity, successfully identifying both the onset and spatial extent of the plug region. This demonstrates the model’s ability to represent regime dependent structural changes in the flow rather than merely scaling velocity magnitude. Even in the strongly viscoplastic regime ($Bi=0.72$), where the plug region occupies most of the channel and only thin shear layers remain near the walls, the model maintains excellent agreement with the reference solution. Notably, RheOFormer preserves both the uniformity of the plug region and the sharp velocity gradients within the boundary layers.

Overall, these results demonstrate that RheOFormer effectively learns the steady state behavior of Bingham fluids across a broad span of viscoplastic behavior, given that training is performed using a large volume of data spanning a broader range than the test conditions. By accurately resolving both shear dominated and plug dominated regimes, the model exhibits strong interpolation capability across the full $Re$–$Bi$ parameter space represented in the training data, providing a reliable baseline for the more challenging generalization studies that follow.

\begin{figure}
\centering
\includegraphics[width=0.45\textwidth]{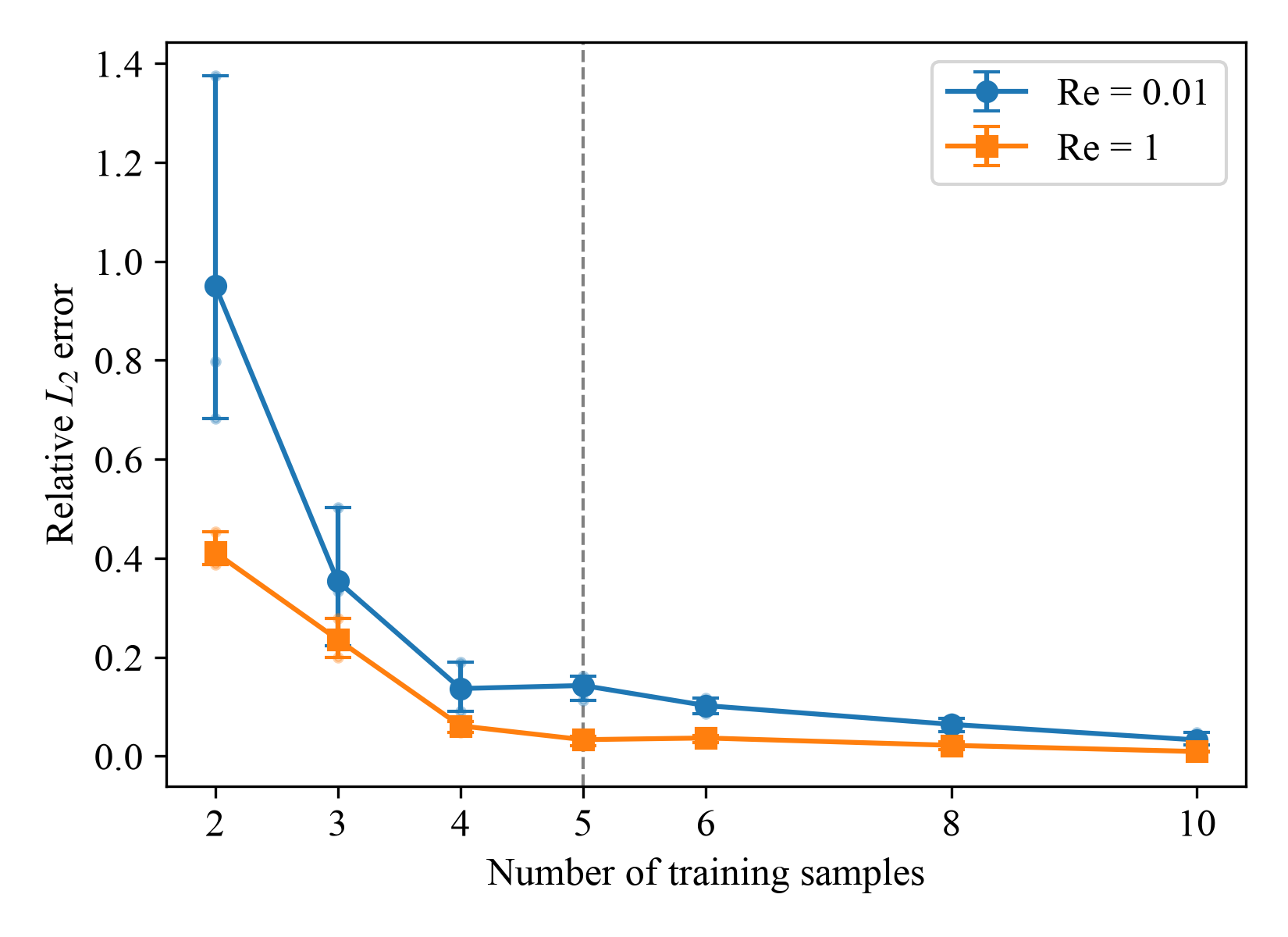}
\caption{Sensitivity of prediction accuracy to the number of training cases, for $Re=0.01$ and $Re=1.0$, over three training seeds. Each sample corresponds to one $(Re,Bi)$ training case selected from the full dataset of 21 $Bi$ per $Re$.}
\label{fig:dataset_size}
\end{figure}

\subsection{Sensitivity to Dataset Size}

Building on the full dataset evaluation, we next systematically test the performance of RheOFormer upon variation of the amount of training data available. Understanding this sensitivity is essential for determining the minimal dataset size required for accurate steady state predictions, particularly since generating high fidelity viscoplastic simulations is computationally costly. To quantify this dependence, we train the model using varying numbers of samples corresponding to different values of $Bi$ and evaluate the resulting prediction error across two representative flow regimes, $Re=0.01$ and $Re=1.0$, which correspond to the lower and upper bounds of the $Re$  range in the dataset.

\begin{figure*}
\centering
\includegraphics[width=\textwidth]{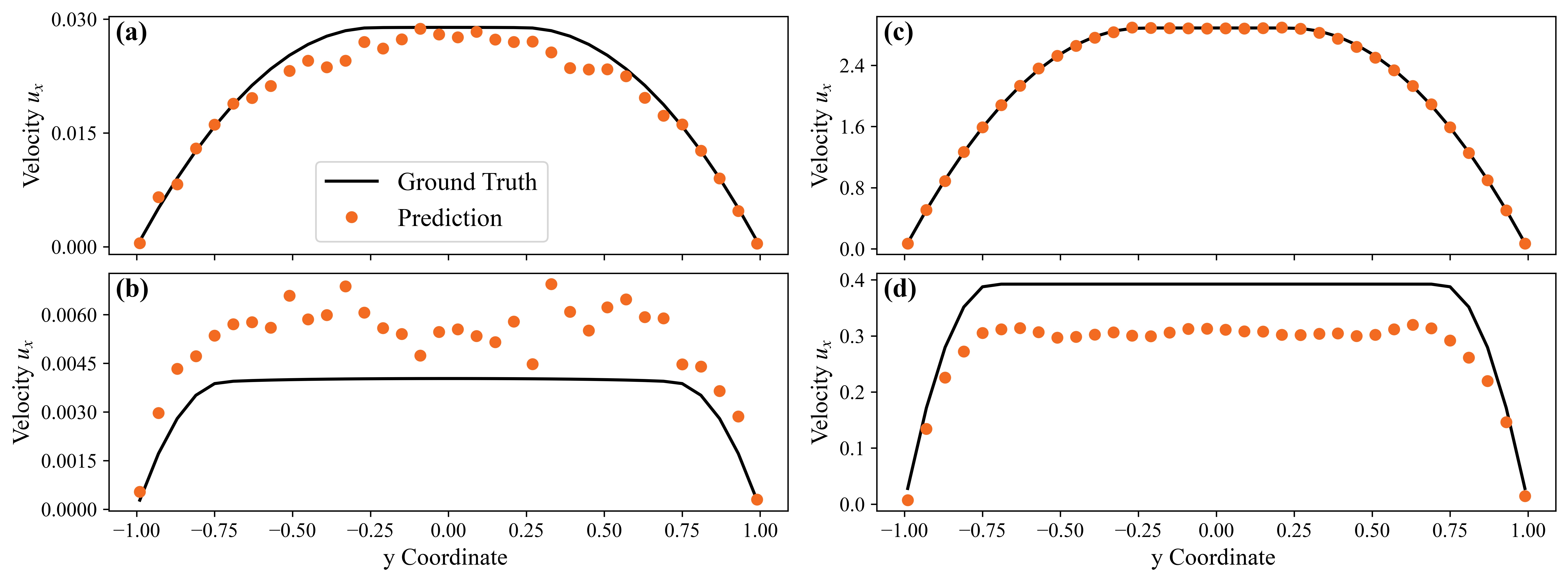}
\caption{Representative steady state velocity profiles across the channel, comparing ground truth (solid black) to RheOFormer predictions (orange dots):
(a) $Re=0.01$, $Bi=0.024$ (mean error $7.2\%$);
(b) $Re=0.01$, $Bi=0.072$ (mean error $46.4\%$);
(c) $Re=1$, $Bi=2.4$ (mean error $0.4\%$);
(d) $Re=1$, $Bi=7.2$ (mean error $21.8\%$).
Panels (a,c) correspond to $Bi/Re = 2.4$ and panels (b,d) to $Bi/Re = 7.2$.
At the lower $Bi/Re$ (a,c), the predictions capture the velocity profile, closely at $Re=1$ and with a slightly underpredicted core at $Re=0.01$.
At the higher $Bi/Re$ (b,d), the model fails to reproduce the flat plug region and systematically underestimates the flatness.}
\label{fig:plateau_failure}
\end{figure*}

Figure~\ref{fig:dataset_size} shows the relative $L_2$ error as a function of the number of samples used during training for two representative flow regimes, $Re=0.01$ and $Re=1.0$. Across both $Re$ values, the error decreases rapidly as the training set expands from 2 to approximately 5 samples. This steep drop indicates that RheOFormer can extract the essential structure of the steady state Bingham flows from only a handful of distinct flow conditions. 
With five training cases, the relative $L_2$ error is reduced to $0.14$ at $Re=0.01$ and $0.03$ at $Re=1$, which corresponds to about 90--95\% of the reduction achieved with ten cases; beyond this point, additional cases yield only diminishing improvements.
These diminishing returns suggest that the global features of the steady state Bingham flow are quickly captured by the model, and that the benefits of adding more training cases are limited by the simplicity and low dimensionality of the geometry. 
Moreover, with the same five cases per $Re$, the log-velocity representation of Section~\ref{Improving} reduces the mean error to about 1\%, indicating that beyond this point accuracy is limited by the representation rather than by the amount of training data.
Motivated by this trend, we adopt five representative cases per $Re$ as the minimal training set size for the in-distribution experiments of Sections~\ref{plateau_error} and~\ref{Improving}; the training sets used in the remaining experiments are specified in the corresponding sections.
This choice strikes a balance between computational efficiency and predictive performance, while still exposing the model to sufficiently diverse rheological behavior.
At $Re = 0.01$ the flow is already steady within the input window ($t \le 0.12$, Section~\ref{sec:temporal}), so that the errors at this $Re$ do not reflect a forecasting difficulty but the model's treatment of small-magnitude fields: a predictor that simply copies the last input would be nearly exact there.

\subsection{Plateau Region Prediction Error} \label{plateau_error}

After establishing the minimal dataset size required for stable training, we next examine RheOFormer’s performance in flow regimes characterized by stronger viscoplastic behavior.
These cases, which feature increasingly pronounced plug regions, provide a stringent test of the model’s ability to resolve low gradient structures that are not emphasized during training.
Figure~\ref{fig:plateau_failure} shows representative velocity profiles for different $Re$ and $Bi$ using only five training samples per $Re$.
While RheOFormer captures the overall boundary layer structure in most cases, a systematic failure mode emerges as the $Bi$ increases.
For low $Bi$ (Figure~\ref{fig:plateau_failure}(a, c)), the predictions remain accurate, with mean errors of 0.4\% at $Re=1$ and 7\% at $Re=0.01$.
As the $Bi$ increases, however, the flow transitions into a plug dominated regime in which $\frac{\partial u_x}{\partial y}$ becomes nearly zero over a substantial portion of the channel.
In these regimes (Figure~\ref{fig:plateau_failure}(b,d)), the model fails to reproduce the plateau velocity and the yield surface correctly.
This behavior occurs at both low and high $Re$ and results in errors of 22\% at $Re=1$ and 46\% at $Re=0.01$, which grow further as $Bi/Re$ approaches its maximum.
At $Re=1$ the shear layers near the walls continue to be well resolved, while the magnitude of the unyielded plug is underpredicted by about 20\% and at $Re=0.01$ the plateau velocity profile becomes oscillatory with high error values.

Since panels (a,c) and (b,d) share the same $Bi/Re$, their ground truth profiles have identical shapes and differ only by a factor of $100$ in magnitude (Section~\ref{sec:data}); the differences in prediction error between them therefore arise from the velocity scale rather than from differences in flow structure.

This failure mode originates from the fact that, in plug dominated flows, the velocity field contains extended regions with extremely small gradients, making the plateau visually and numerically subtle. When trained with the minimal dataset, the model naturally prioritizes fitting the high gradient boundary layers (where most velocity variation occurs) but pays insufficient attention to the low gradient regions that define the plug. This observation motivates the approaches investigated in Section \ref{Improving}, where we introduce additional input features and loss formulations specifically designed to amplify or explicitly target the low gradient regions responsible for the plateau.

\begin{figure*}
\centering
\includegraphics[width=\textwidth]{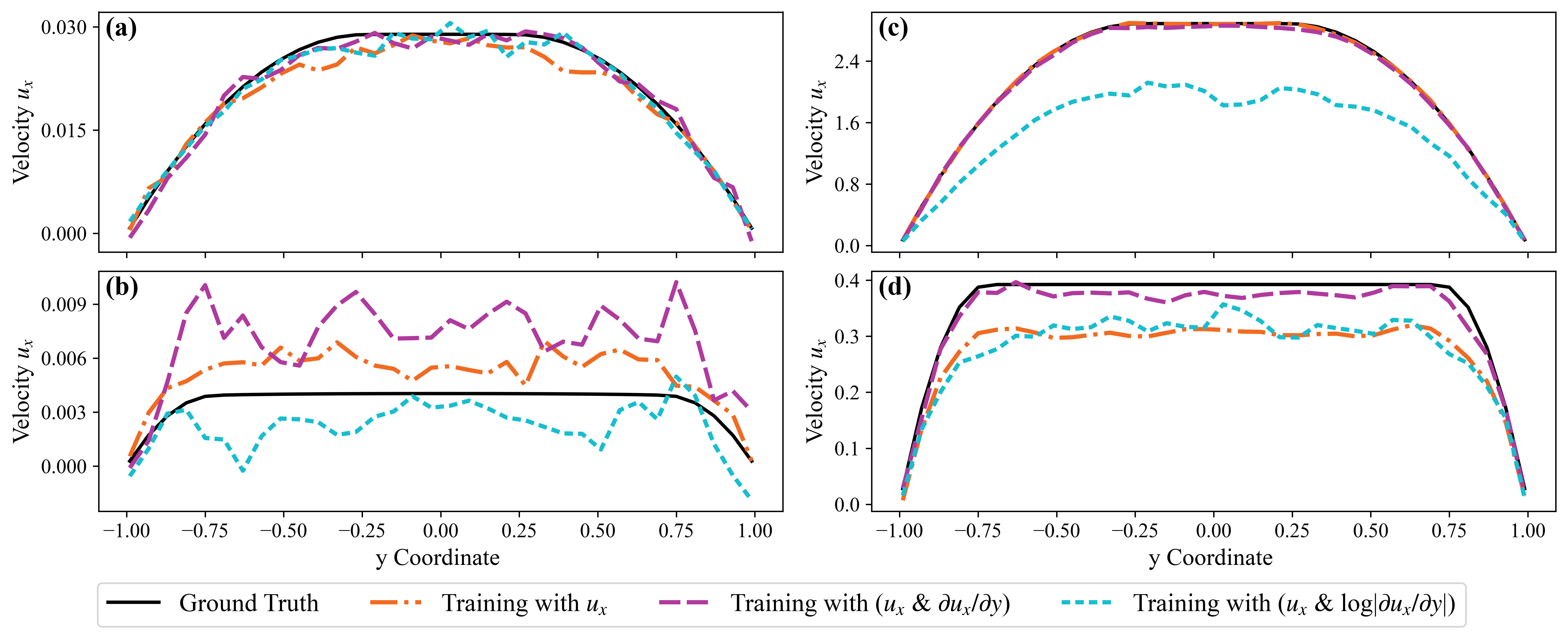}
\caption{Comparison of input representations for predicting the plateau velocity, all trained with the relative $L_2$ loss.
Each panel compares the ground truth (solid black) with RheOFormer predictions obtained with (i) the velocity $u_x$ alone (orange, dash-dotted), (ii) $u_x$ augmented with $\partial u_x/\partial y$ (magenta, dashed), and (iii) $u_x$ augmented with $\log|\partial u_x/\partial y|$ (cyan, dotted).
The left column corresponds to $Re=0.01$ and the right column to $Re=1$: 
(a) $Bi=0.024$;
(b) $Bi=0.072$;
(c) $Bi=2.4$;
(d) $Bi=7.2$.
At $Re=1$ (c,d), the $\partial u_x/\partial y$ input reproduces the profile to within about 3\% and recovers the plateau height that the velocity-only model underpredicts, reducing the error from 21.8\% to 4.3\%, whereas the log-gradient input is less reliable, underpredicting the profile and not improving the plateau.
At $Re=0.01$ (a,b), the $\partial u_x/\partial y$ input introduces strong oscillations and overpredicts the plateau in (b), and the log-gradient input underpredicts it with oscillations and unphysical negative velocities near the walls.}
\label{fig:plateau_improvements}
\end{figure*}

\subsection{Improving Prediction Accuracy} \label{Improving}

Section \ref{plateau_error} revealed a systematic failure mode of RheOFormer when predicting flows with pronounced plug regions: the model fails to reproduce the plateau velocity and hence fails to reproduce the unyielded region of the flow at high $Bi$, particularly when extended plug regions dominate the flow. To address this limitation, we investigate several modifications to the input representation and loss formulation, with the goal of making plug region information more prominent during training. 

Figure~\ref{fig:plateau_improvements} summarizes the performance of the input-based strategies across representative $Re$--$Bi$ cases.
Each panel compares the ground truth solution with predictions obtained using three different input configurations, all trained with the relative $L_2$ loss: (i) the raw velocity field $u_x$; (ii) the velocity field $u_x$ augmented with its gradient $\frac{\partial u_x}{\partial y}$; and (iii) the velocity field $u_x$ augmented with the logarithm of the gradient $\log(|\frac{\partial u_x}{\partial y}|)$.
These configurations are designed to progressively increase the model’s sensitivity to regions of weak shear by explicitly exposing gradient information and amplifying small magnitude variations. The following subsections analyze each approach in detail.\\

\textbf{Augmenting Inputs with Velocity Gradients:} The first modification  we explore is augmenting the input with the gradient field $\frac{\partial u_x}{\partial y}(x,y,t)$. The motivation for this approach is that unyielded plug regions are characterized by (nearly) vanishing velocity gradients, whereas yielded regions exhibit finite shear rates. By supplying $\frac{\partial u_x}{\partial y}(x,y,t)$ as an additional input channel, the attention mechanism can exploit these cross-channel contrasts to more accurately infer the yield surfaces, thereby improving its ability to distinguish between plug regions and adjacent shear layers and to reconstruct the plateau region.

As seen in Figure~\ref{fig:plateau_improvements}, adding $\frac{\partial u_x}{\partial y}$ improves performance for cases with moderately large $Bi$, specifically at $Re=1$ (panels~(c) and (d)) where the predicted profile and plateau height fall within about 3\% of the ground truth; for the plug-dominated case (d), the error decreases about $5$ times compared to training only with $u_x$.
This indicates that explicit access to shear rate information can enhance the model’s ability to infer the extent of the plug region when the velocity and its gradient are sufficiently large.
However, in the strongly viscoplastic regimes (panels~(b) and (d)), the limitations of this approach become clear.
Because the gradient is nearly zero across most of the domain in plug dominated flows, the added feature provides little meaningful signal and is easily dominated by numerical noise.
This issue is most severe in low $Re$  cases, where both the velocity magnitudes and their gradients are extremely small; in such regimes, the numerical noise becomes comparable to the signal itself, leading to visibly degraded predictions.
As a result, the gradient-augmented model fails to reproduce the flat central region accurately. These observations indicate that simply providing raw gradient information is still insufficient for reliably resolving strongly plug dominated flows.\\

\textbf{Augmenting Inputs with Logarithm of Velocity Gradients:} To better resolve the near zero gradient values inside the plug, we next  introduce the logarithm of the gradient  magnitude, $\log(|\frac{\partial u_x}{\partial y}|)$, as an additional input channel. This transformation expands the dynamic range of values that cluster near zero and amplifies subtle variations associated with the yielding boundary. Unlike the raw gradient, the log-gradient provides meaningful structure  throughout the domain, even when $\frac{\partial u_x}{\partial y}$ is extremely small, making it more informative for learning the plug region behavior.
This expectation is borne out at $Re=0.01$ (Figure~\ref{fig:plateau_improvements}(a,b)), where the log-gradient input gives the most accurate predictions of the three representations, with errors of 4--6\% in (a) and 36--45\% in (b) across training seeds, although the plateau is still underpredicted and the profiles remain oscillatory.
At $Re=1$, however, the log-gradient input does not improve upon the raw gradient and its predictions vary strongly between training seeds, with errors of 4--31\% in (c) and 6--21\% in (d).
A plausible reason is the short input window ($t\le0.12$): at high $Re$ the flow in the channel core has not yet developed at these times, so that the near-zero early-time gradients are strongly amplified by the logarithm and dominate the input, whereas at $Re=0.01$ the flow develops much faster.
Among gradient-based inputs, the raw gradient therefore provides the largest gains at high $Re$, while the log-gradient is advantageous only at low $Re$, and neither fully resolves the plateau in strongly plug-dominated, low-$Re$ flows. Averaged over all cases, the mean error is 6.3\% with the raw gradient and 13.7\% with the log-gradient, compared with 5.9\% for $u_x$ alone.\\

\begin{figure*}[t]
\centering
\includegraphics[width=\textwidth]{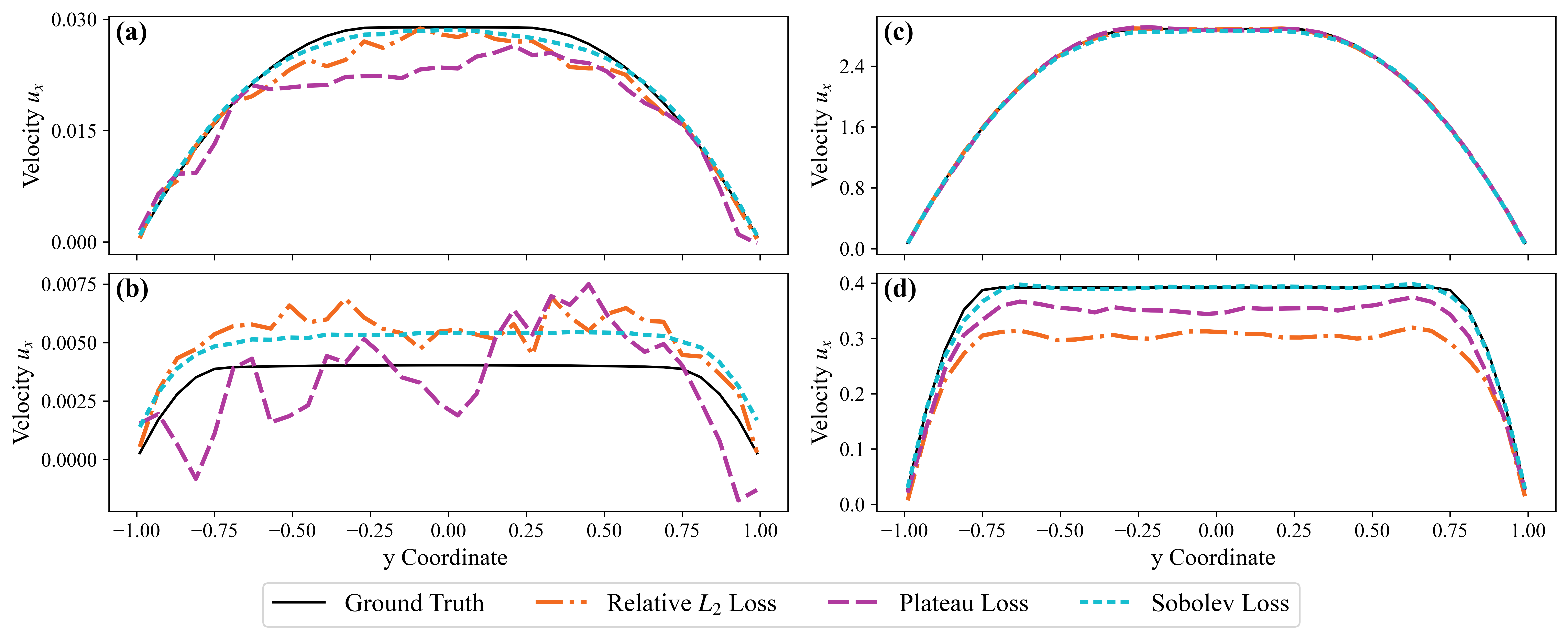}
\caption{Effect of different loss formulations on prediction accuracy for representative $Re$--$Bi$ cases. 
Each panel compares the ground truth velocity profile (solid black) with predictions obtained using (i) the baseline relative $L_2$ loss (orange, dash-dotted), (ii) a plateau-weighted loss (magenta, dashed), and (iii) a Sobolev loss (cyan, dotted). The left column corresponds to $Re=0.01$ and the right column to $Re=1$:
(a) $Bi=0.024$; (b) $Bi=0.072$;
(c) $Bi=2.4$;
(d) $Bi=7.2$.
At $Re=1$ both modified losses raise the plateau height that the baseline underpredicts, with the Sobolev loss performing best.
At $Re=0.01$, the plateau loss introduces large oscillations and unphysical negative velocities at the walls, while the Sobolev loss yields smooth profiles with the lowest error of the three.}
\label{fig:loss_comparison}
\end{figure*}

\textbf{Loss Formulations for Low Shear Emphasis:} In addition to modifying the input representation, we evaluated several loss formulations designed to emphasize the plateau region directly. Three losses were tested: (i) the baseline relative $L_2$ loss, (ii) a plateau-weighted loss that amplifies errors at grid points where $|\nabla u_{\mathrm{gt}}| \le \tau_{\mathrm{th}}$ (Eq.~\ref{eq:mask}), and (iii) a Sobolev loss combining velocity and gradient errors.
The baseline uses $u_x$ as the only input, while the plateau-weighted and Sobolev models receive $u_x$ together with $\partial u_x/\partial y$, which is also needed to evaluate these two objectives.
Figure~\ref{fig:loss_comparison} summarizes the performance of these losses across representative $(Re,Bi)$ cases.

The relative $L_2$ loss performs well when the velocity field contains significant spatial variation (e.g.\ low $Bi$ flows), but it fails to reproduce the plateau velocity for large $Bi$, where the gradients are nearly zero throughout most of the domain. Because the $L_2$ norm naturally focuses on high magnitude variations, it allocates disproportionate weight to the boundary layers and fails to sufficiently penalize errors in the plug. As a result, the model trained with this loss systematically underweights the plateau and fails to reproduce the flat central profile. The plateau-weighted loss attempts to correct this imbalance by explicitly increasing the contribution of low shear regions to the objective function.
While this strategy improves accuracy in high $Re$ cases, it becomes highly sensitive to the numerical values of the velocity field and tends to overlook small magnitudes and their variations during training, leading to instability and degraded predictions at low $Re$ (panels~(a) and (b)). Consequently, its performance varies significantly across the parameter space, producing large errors in some regimes despite improving others.

The Sobolev loss provides a more balanced and physically grounded alternative.
By penalizing discrepancies in both the velocity and its spatial derivatives, it encourages the model to capture the flatness of the plug and the sharpness of the shear layers simultaneously.
As shown in Figure~\ref{fig:loss_comparison}, the Sobolev loss produces smooth, oscillation-free profiles and provides consistently improved predictions across all cases and training seeds, reducing the errors in (a) and (d) from 7--22\% to 1--3\% without requiring case specific tuning.
Averaged over all cases, the mean error is 3.0\% with the Sobolev loss, compared with 5.9\% for the baseline and 7.3\% for the plateau-weighted loss.
Among the loss-based strategies evaluated, the Sobolev loss delivers the most stable and broadly applicable enhancement to the model’s accuracy, which is also physically meaningful, whereas the plateau-weighted loss gives smaller gains than the Sobolev loss at $Re=1$ and is strongly unstable at $Re=0.01$.\\

\textbf{Logarithmic Velocity Representation:} Finally, we investigate representing both the input and output velocity fields in logarithmic space, i.e., mapping $u_x \mapsto \log|u_x|$ before passing the data to the model. Since differences in the plateau region are small in absolute magnitude, this transformation rescales low velocity variations, amplifying subtle differences and enabling the model to identify patterns that are otherwise obscured.

\begin{figure*}[t]
\centering
\includegraphics[width=\textwidth]{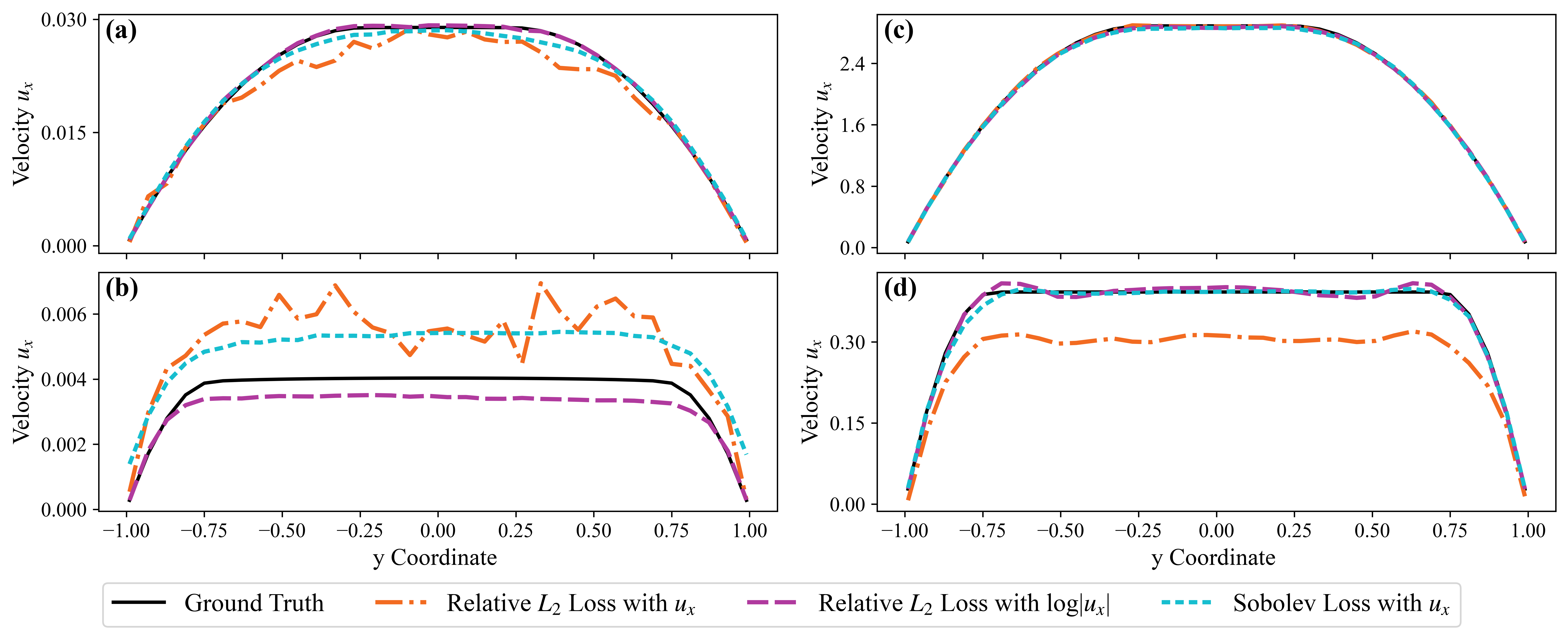}
\caption{Comparison of the most effective input and loss strategies for capturing the plug-flow plateau.
Each panel compares the ground truth velocity profile (solid black) with RheOFormer predictions obtained with (i) the relative $L_2$ loss on $u_x$ (orange, dash-dotted), (ii) the relative $L_2$ loss on $\log|u_x|$ (magenta, dashed), and (iii) the Sobolev loss (cyan, dotted).
The left column corresponds to $Re=0.01$ and the right column to $Re=1$:
(a) $Bi=0.024$;
(b) $Bi=0.072$;
(c) $Bi=2.4$;
(d) $Bi=7.2$.
At $Re=1$ both log-velocity training and the Sobolev loss reproduce the plateau height to within about 2\%, clearly outperforming the baseline.
At $Re=0.01$, all three approaches show plateau errors, while log-velocity training gives the most accurate predictions in both cases.
In the plug-dominated case (b), log-velocity training yields a smooth profile with an underpredicted plateau, while the Sobolev loss overpredicts the plateau (38.1\%) and the baseline overpredicts it with large oscillations (46.4\%).}
\label{fig:logV}
\end{figure*}

This representation leads to a substantial improvement in the predictive accuracy at larger velocities (e.g. $Re=1$). Figure~\ref{fig:logV} illustrates the effect of this transformation across a range of $(Re,Bi)$ cases. 
In each panel, the log-velocity model is compared directly against the velocity-only baseline and the Sobolev loss. 
At $Re=1$, log-scaled training performs comparably to the Sobolev loss and much better than the baseline. 
Even in the most challenging high $Bi$ regimes, the log-velocity model achieves errors as low as 2\%, accurately reproducing both the plug height and the shear layer transitions. Averaged over all cases, log-velocity training gives the lowest mean error of all strategies considered (1.0\%).
A key advantage of the log-velocity representation is its numerical stability. 
Unlike gradient-based inputs, which can be sensitive to noise when gradients are extremely small, the log-transform remains smooth and well behaved across the entire domain. 
At $Re=0.01$, although the log-velocity prediction remains smooth and free of oscillations, it still shows a small discrepancy for the velocity magnitudes, indicating that the low-$Re$ plateau remains challenging even with this representation.\\

These findings show that the most robust and broadly effective improvements arise from the use of log-scaled velocities and Sobolev loss, both of which amplify the subtle variations that define the plug region, with the clearest gains at larger velocities. 
The consistency of these enhancements across all tested flow regimes indicates that the plateau region prediction error is not fundamentally tied to model capacity or data scarcity, but rather to the numerical characteristics inherent to viscoplastic steady states, specifically, the prevalence of extremely small gradients that obscure meaningful structure. 
By adopting input representations that better emphasize low shear regions, RheOFormer is able to recover accurate predictions even in strongly viscoplastic flows with large plug regions.

\begin{figure*}[t]
\centering
\includegraphics[width=\textwidth]{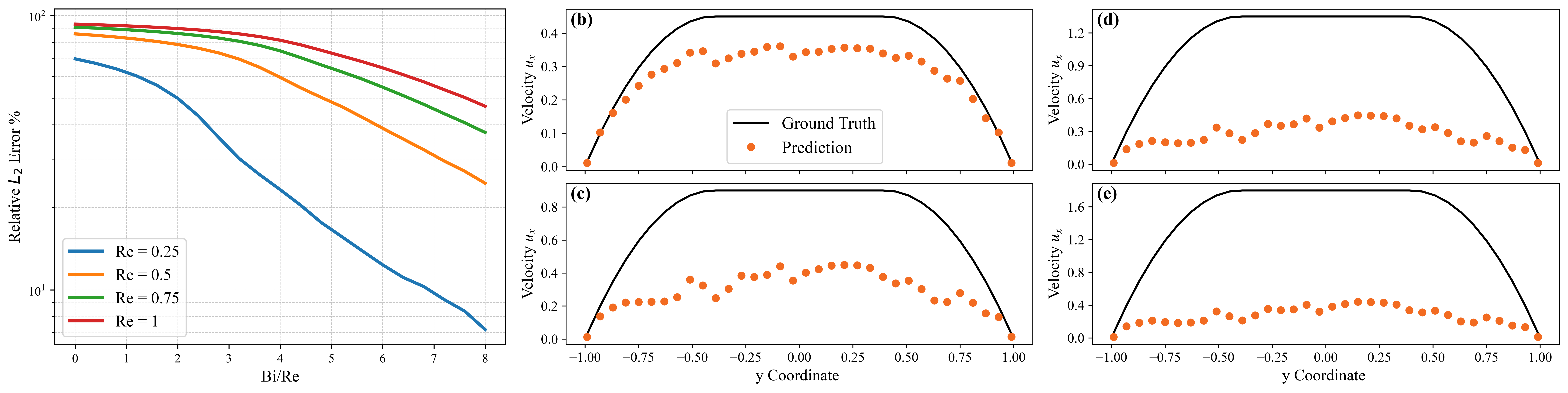}
\caption{Extrapolation performance of the model trained only on low Reynolds numbers ($Re = 0.01, 0.025, 0.05, 0.075, 0.1$) and evaluated on higher $Re$ flows. 
(a) Mean prediction error as a function of $Bi/Re$ for $Re = 0.25, 0.5, 0.75,$ and $1.0$. 
(b) $Re = 0.25$, 
(c) $Re = 0.5$, 
(d) $Re = 0.75$, and 
(e) $Re = 1.0$: representative velocity profiles, each at the $Bi$ near the midpoint of its range.}
\label{fig:lowRe_extrap}
\end{figure*}

\subsection{Extrapolation Across the $Re$ Range} \label{Re_extrapolation}

\begin{figure*}[t]
\centering
\includegraphics[width=\textwidth]{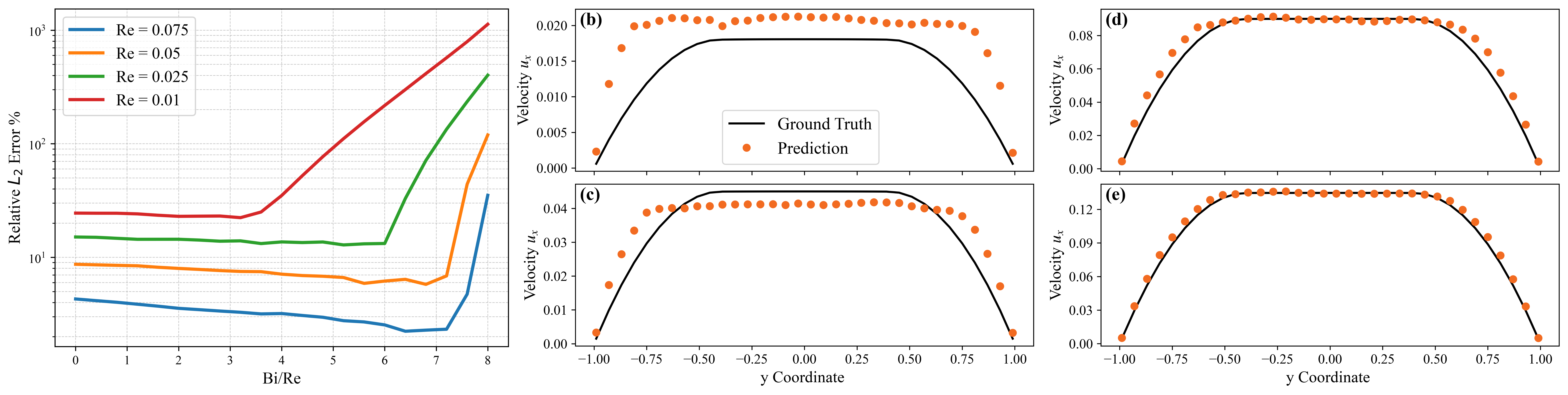}
\caption{Extrapolation performance of the model trained only on high Reynolds numbers ($Re = 0.1, 0.25, 0.5, 0.75, 1.0$) and evaluated on lower $Re$ flows. (a) Mean prediction error across the $Bi/Re$ range for $Re = 0.01, 0.025, 0.05,$ and $0.075$. 
(b) $Re = 0.01$, 
(c) $Re = 0.025$, 
(d) $Re = 0.05$, and 
(e) $Re = 0.075$: representative velocity profiles, each at a characteristic $Bi$.}
\label{fig:highRe_extrap}
\end{figure*}

The preceding sections focused on improving accuracy within the regime covered by the training dataset, particularly for flows exhibiting strong plug formation. We now examine a different and complementary challenge: the ability of RheOFormer to extrapolate across the $Re$ dimension.
Since the steady profile scales linearly with $Re$ at fixed $Bi/Re$ (Section~\ref{sec:data}), extrapolation along $Re$ probes the model's ability to predict velocity magnitudes outside the training range rather than new flow structures.
In many operator learning settings, interpolation within the training distribution is achievable, whereas true extrapolation remains fundamentally difficult. To quantify these effects, we evaluate two extrapolation scenarios: (i) models trained exclusively on low $Re$ and tested on higher $Re$, and (ii) models trained on high $Re$ and tested at lower $Re$. Figures~\ref{fig:lowRe_extrap} and \ref{fig:highRe_extrap} summarize the prediction errors and representative velocity profiles for these two settings.
Unless otherwise stated, all extrapolation experiments in Sections~\ref{Re_extrapolation}--\ref{resolve} use the log-velocity configuration of Section~\ref{Improving}, the most accurate configuration identified above, so that the effect of parameter-space coverage is isolated from that of the representation and loss choices examined in previous Sections.\\

\textbf{Extrapolation When Training on Low $Re$:}
We first train RheOFormer using only the flow profiles in the low Reynolds ($Re = 0.01, 0.025, 0.05, 0.075, 0.1$) regime, using all 21 $Bi$ values at each $Re$ (105 training cases), and test it on flows at higher $Re$. 
This test directly probes whether a model exposed exclusively to low magnitude velocity fields can extrapolate to larger velocities associated with higher $Re$. Figure~\ref{fig:lowRe_extrap}(a) reports the mean prediction error across the full $Bi$ range for test cases with $Re=0.25,\,0.5,\,0.75,$ and $1.0$. Two clear trends emerge. First, the prediction error increases systematically as the $Re$  grows. The model performs reasonably well for some cases with $Re=0.25$, which lies close to the training regime, but accuracy degrades sharply as $Re$ moves further from the values seen during training. For $Re=0.75$ and $1.0$, the errors exceed 35\% over the entire $Bi$ range, indicating that the model cannot reconstruct the substantially larger velocity magnitudes associated with higher $Re$ flows. Second, the error decreases with increasing $Bi$. This behavior is expected since higher $Bi$ suppresses the maximum velocity and yields profiles that are closer in magnitude to those encountered in the low $Re$ training set.

Panels~(b)–(e) in Figure~\ref{fig:lowRe_extrap} show representative predictions, using for each $Re$ the case whose $Bi$ lies near the midpoint of its testing range. At $Re=0.25$ [Figure~\ref{fig:lowRe_extrap}(b)], the model captures the overall shape of the velocity field with mean errors near $20\%$, although deviations remain noticeable, specifically in the plateau. As the $Re$  increases, however, the model fails to reproduce the correct velocity magnitude. In Figures~\ref{fig:lowRe_extrap}(c)–(e), corresponding to $Re=0.5$, $0.75$, and $1.0$, the predicted profiles saturate around $u_x \approx 0.45$, far below the true peak values. This occurs because the largest velocities observed during training correspond to $Re=0.1$, leaving the model without exposure to the higher magnitudes required to reconstruct the target solutions.

These observations demonstrate that training exclusively on low $Re$ regime enables only limited upward extrapolation. While the model can interpolate slightly beyond the training regime, its predictive accuracy rapidly collapses as the characteristic velocity scale departs from the range encoded in the training data. This highlights the importance of sampling across the full dynamical range when training a single operator intended for broad parameter coverage.\\

\textbf{Extrapolation When Training on High $Re$:}
We now examine the inverse scenario to the previous subsection by training RheOFormer exclusively on high $Re$ regime ($Re = 0.1, 0.25, 0.5, 0.75, 1.0$; all 21 $Bi$ values at each $Re$, 105 training cases) and evaluating its performance on low $Re$ flows. This setting tests whether a model exposed primarily to large velocity magnitudes can recover the lower speed profiles characteristic of small $Re$ flows. Figure~\ref{fig:highRe_extrap}(a) reveals a contrasting 
trend compared to the previous subsection. When predicting flows down to approximately $Re\approx 0.05$–$0.075$, the model maintains moderate error levels over a broad range of $Bi$. This behavior reflects the fact that training on high $Re$ cases exposes the model not only to large velocity magnitudes at low $Bi$, but also to moderately small velocities at higher $Bi$ values. As $Bi$ decreases, the velocity magnitude increases, bringing the target profiles closer to those values already encountered during training. Consequently, the model is able to represent nearby low $Re$ flows with reasonable quantitative accuracy. Nonetheless, the errors begin to rise sharply in cases where the velocity magnitudes fall below the range represented in the training data. Even for $Re=0.075$, high $Bi$ flows exhibit a significant increase in error, indicating that the model no longer has sufficient basis functions to reconstruct the solution accurately. This behavior mirrors the limitations observed in the previous subsection: once the target field departs substantially from the dynamical range encoded during training, performance rapidly degrades.

Panels~(b)–(e) of Figure~\ref{fig:highRe_extrap} show representative predictions at characteristic $Bi$ for each $Re$. For $Re=0.075$ [Figure~\ref{fig:highRe_extrap}(e)], the model performs well with errors near $4\%$. At intermediate $Re$ of $0.025$ and $0.05$ [panels~(c) and (d)], the predictions remain  qualitatively accurate, though errors increase to the $8$–$18\%$ range. By contrast, when extrapolating to $Re=0.01$ [panel~(b)], the model fails to capture the correct velocity magnitude and produces errors close to $50\%$, primarily due to systematic overestimation of the plug height. These large discrepancies arise because such low velocity fields are absent from the training regime, preventing the model from reconstructing the required behavior. An asymmetry emerges when comparing the two extrapolation directions. Models trained on high $Re$ flows extrapolate downward more effectively than models trained on low $Re$ flows extrapolate upward.
Since the steady profile shapes at fixed $Bi/Re$ are identical across $Re$, this asymmetry reflects the range of velocity magnitudes contained in the training data rather than differences in flow structure: high-$Re$ training sets already include moderately small velocities at high $Bi$, whereas no low-$Re$ case reaches the velocity magnitudes of high-$Re$ flows.
Nevertheless, both extrapolation directions ultimately fail when the target $Re$  lies outside the dynamical regime captured during training.

Taken together, these results demonstrate that RheOFormer, like most operator learning frameworks, performs reliably within the convex hull of its training distribution but degrades sharply when asked to extrapolate across dynamically distinct regions of the $Re$–$Bi$ parameter space. This motivates the investigation in Section \ref{resolve}, where we show that strategically sampling representative cases across the full parameter range substantially reduces these extrapolation failures.

\begin{figure*}[t]
\centering
\includegraphics[width=\textwidth]{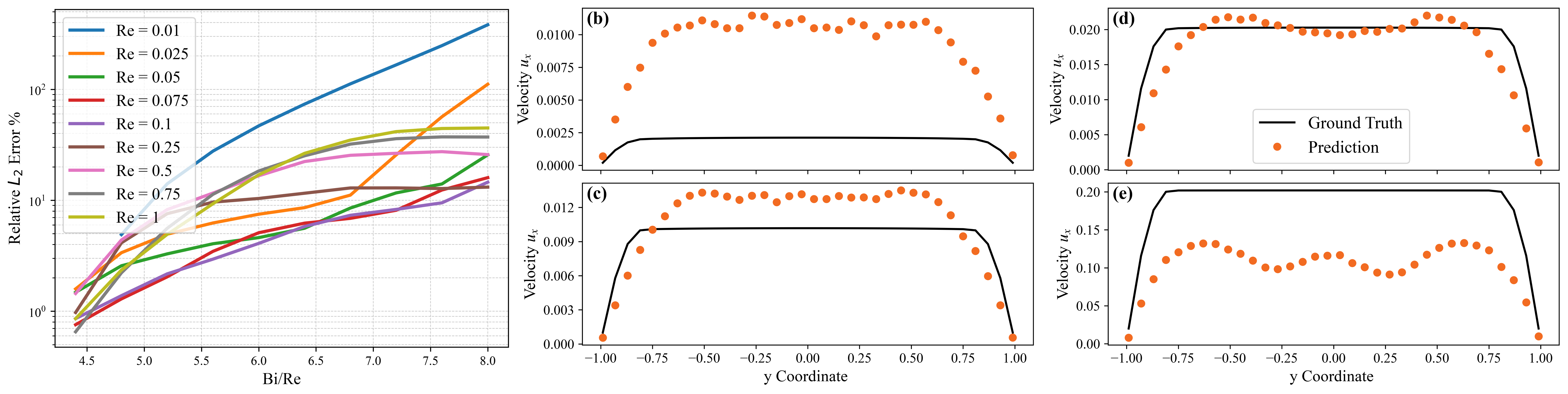}
\caption{Extrapolation performance of RheOFormer trained only on low $Bi$ ($Bi/Re \le 4$) and evaluated on flows with larger $Bi$. 
(a) Mean prediction error as a function of $Bi/Re$ over the testing range.
% (b)--(e) Representative velocity profiles at the largest tested $Bi$: 
% (b) $Re = 0.01$, $Bi = \text{XX}$; (c) $Re = 0.05$, $Bi = \text{XX}$; (d) $Re = 1.0$, $Bi = \text{XX}$; (e) $Re = 0.1$, $Bi = \text{XX}$.}}
(b) $Re = 0.01$, 
(c) $Re = 0.05$, 
(d) $Re = 0.1$, and 
(e) $Re = 1.0$: representative velocity profiles at the largest tested $Bi$.}
\label{fig:lowBi_extrap}
\end{figure*}

\subsection{Extrapolation Across the $Bi$ Range}
Having examined the extrapolation capabilities along $Re$ axis, we now turn to the second key control parameter in Bingham flows: the $Bi$. Whereas variations in $Re$ only rescale the steady velocity magnitude, changes in $Bi$ control the strength and spatial extent of the plug region through the yield stress. 
As a result, varying $Bi$ induces qualitative changes in the flow structure rather than a simple rescaling of velocity magnitude. 
Extrapolating across $Bi$ is therefore a stringent test of whether RheOFormer has learned the underlying physics rather than simply interpolating among a few observed plug shapes.\\

\textbf{Extrapolation When Training on Low $Bi$:}\label{low_Bi} We first consider models trained only on weakly viscoplastic flows, i.e.\ on the lower half of the $Bi$ range, in which the plug occupies at most $40\%$ of the channel, and then evaluate the performance on progressively larger $Bi$ where pronounced plug regions develop. In this setting, the training set contains velocity profiles with relatively mild flattening, and the model is asked to extrapolate toward increasingly plug dominated regimes that lie outside the observed range.

Here, the $Bi$ domain for each $Re$ is divided into two equal halves. The lower half of the $Bi$ range, corresponding to weakly viscoplastic flows near the Newtonian limit, is used for training, while the upper half with stronger viscoplastic behavior is reserved for testing. 
In Figure~\ref{fig:lowBi_extrap}(a), $Bi$ is normalized by $Re$, so that the parameter range spans $0 \le Bi/Re \le 8$ for every $Re$; the testing range therefore corresponds to $Bi/Re \ge 4$, and the left edge of the axis marks the boundary with the training range.
This normalization provides a consistent reference point for comparing the extrapolation behavior across all $Re$ values. 
Figure~\ref{fig:lowBi_extrap} summarizes the resulting behavior. 
Panel~(a) shows the mean prediction error as a function of the $Bi$ range for several $Re$. 
For all $Re$ values, the prediction error generally increases as $Bi$ moves farther into the testing range, indicating increasing difficulty in extrapolating toward stronger viscoplastic behavior. 
The magnitude of this degradation, however, depends strongly on $Re$. 
The largest errors occur at low $Re$, with $Re=0.01$ showing an increase of about two orders of magnitude across the extrapolation range. 
This suggests that the model, having been trained primarily on weakly viscoplastic profiles near the Newtonian limit, has limited ability to extrapolate the development of plug-dominated flow when both $Bi$ is high and $Re$ is low.
In contrast, the error increase is considerably smaller at moderate $Re$ ($0.075$--$0.25$), where the errors remain below about 12\%. At the highest $Re$ ($0.5$--$1$), however, the errors again grow to 25--45\% at the largest $Bi$.

Panels~(b)–(e) of Figure~\ref{fig:lowBi_extrap} illustrate representative predictions at the largest $Bi$ tested for each $Re$. 
At $Re=0.01$ and $Re = 0.05$ [Figure~\ref{fig:lowBi_extrap}(b), (c)], the predictions are qualitatively incorrect: the model overestimates the velocity magnitude, produces a strongly distorted profile with spurious oscillations, and completely fails to represent the plateau region, resulting in extremely large errors. 
For $Re=0.1$ [panel~(d)], the model recovers the overall shape of the velocity profile but still underestimates the degree of flattening in the plateau region and the slope near the yield surfaces, resulting in an error of $11\%$. 
At $Re=1.0$ [panel~(e)] however, the plateau velocity is underestimated by about 40\% and the profile contains large oscillations. 
In all cases, the plateau is not correctly reproduced because the model has never observed plug-like behavior in the training data and therefore lacks the functional structure needed to reconstruct it.
These results indicate that training solely on low $Bi$ flows does not provide sufficient information for the model to infer how the plug region develops as the yield stress increases. 
The operator learned in this regime primarily encodes Newtonian-like behavior and fails to extrapolate to strongly viscoplastic states.\\

\begin{figure*}[t]
\centering
\includegraphics[width=\textwidth]{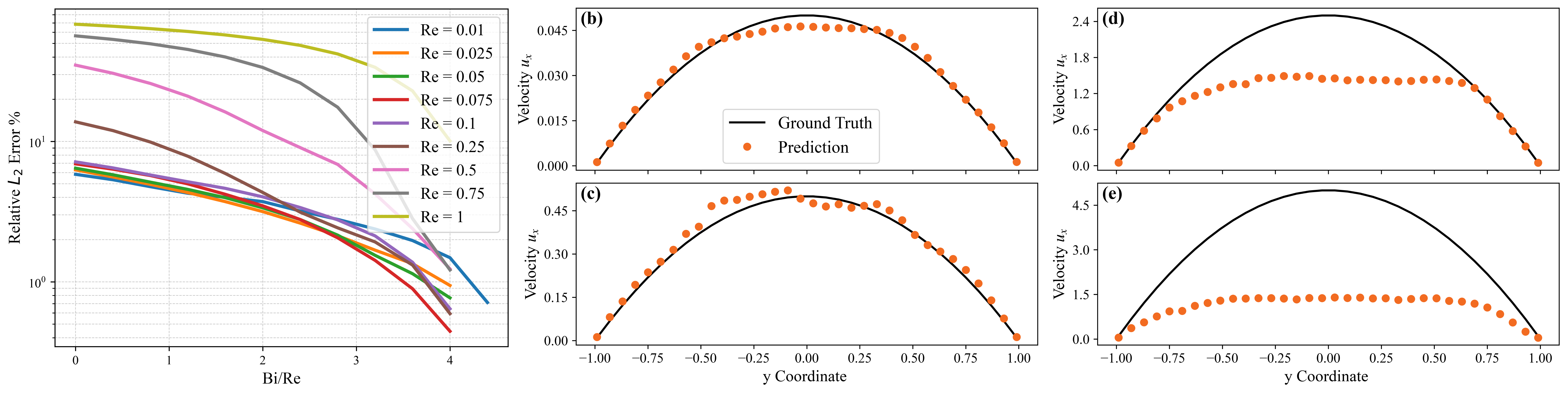}
\caption{Extrapolation performance of RheOFormer trained only on large $Bi$ and evaluated on flows with smaller $Bi$. 
(a) Mean prediction error as a function of $Bi/Re$ over the testing range.
(b) $Re = 0.01$, 
(c) $Re = 0.1$, 
(d) $Re = 0.5$, and 
(e) $Re = 1.0$: representative velocity profiles at the Newtonian limit ($Bi = 0$).}
\label{fig:highBi_extrap}
\end{figure*}

\textbf{Extrapolation When Training on High $Bi$:} To complement the preceding analysis, we next consider the inverse extrapolation scenario in which RheOFormer is trained solely on strongly viscoplastic flows with large $Bi$. 
In this training regime, the velocity fields are dominated by pronounced plug regions with minimal curvature. The model is then evaluated on flows with progressively smaller $Bi$, where the velocity profiles transition toward more curved, Newtonian-like shapes. 
This test probes whether a model trained only on flattened, plug dominated profiles can extrapolate toward the fully yielded regime. For each $Re$, the upper half of the $Bi$ range is used for training, while the lower half is reserved for testing (the reverse of the low-$Bi$ setup above). 
In Figure~\ref{fig:highBi_extrap}(a), the testing range corresponds to $0 \le Bi/Re \lesssim 4$; the right edge of the axis marks the boundary with the training range, and the extrapolation distance increases toward $Bi/Re = 0$.

Figure~\ref{fig:highBi_extrap}(a) shows that the prediction error generally increases as the test $Bi$ moves farther from the training regime. 
For low and moderate $Re$, this increase is gradual, indicating that the model can partially adjust to the weakening of the plug region. However, for higher $Re$ values ($0.25$--$1.0$), the errors are substantially larger. This deterioration is expected, as the model has never encountered high curvature profiles or large velocity magnitudes in the high $Bi$ training set. Panels~(b)–(e) show representative predictions at the Newtonian limit for each $Re$. 
At $Re=0.01$ and $Re=0.1$ (Figures~\ref{fig:highBi_extrap}(b, c)), the model produces smooth profiles with modest errors around 6\%, but retains a residual plateau and fails to recover the correct parabolic curvature. 
This reflects the mismatch between the training data where all examples contain a plug, and the Newtonian limit where no such plateau exists. 
At intermediate $Re$ of $0.50$ [panel~(d)], the model increasingly underestimates centerline curvature and magnitude, leading to errors of roughly $25\%$. 
At the highest $Re$ of $1.0$, the predictions deteriorate sharply: the model cannot reproduce the large velocity values or the steep gradients in the shear layer and instead outputs an overly flattened profile with errors of about $65\%$ [panel~(e)].

These results reveal an extrapolation asymmetry complementary to that observed when training on low $Bi$. Whereas extrapolation toward higher $Bi$ fails because the model has not learned the formation of plug regions, extrapolation toward lower $Bi$ fails because the model lacks exposure to the high curvature and large shear rates characteristic of Newtonian-like flow. Both directions underscore the necessity of broad parameter-space sampling when training a single neural operator intended to generalize across viscoplastic regimes.

\subsection{Resolving Extrapolation Limitations by Targeted Sampling} \label{resolve}

\begin{figure}
    \centering
    \includegraphics[width=\linewidth]{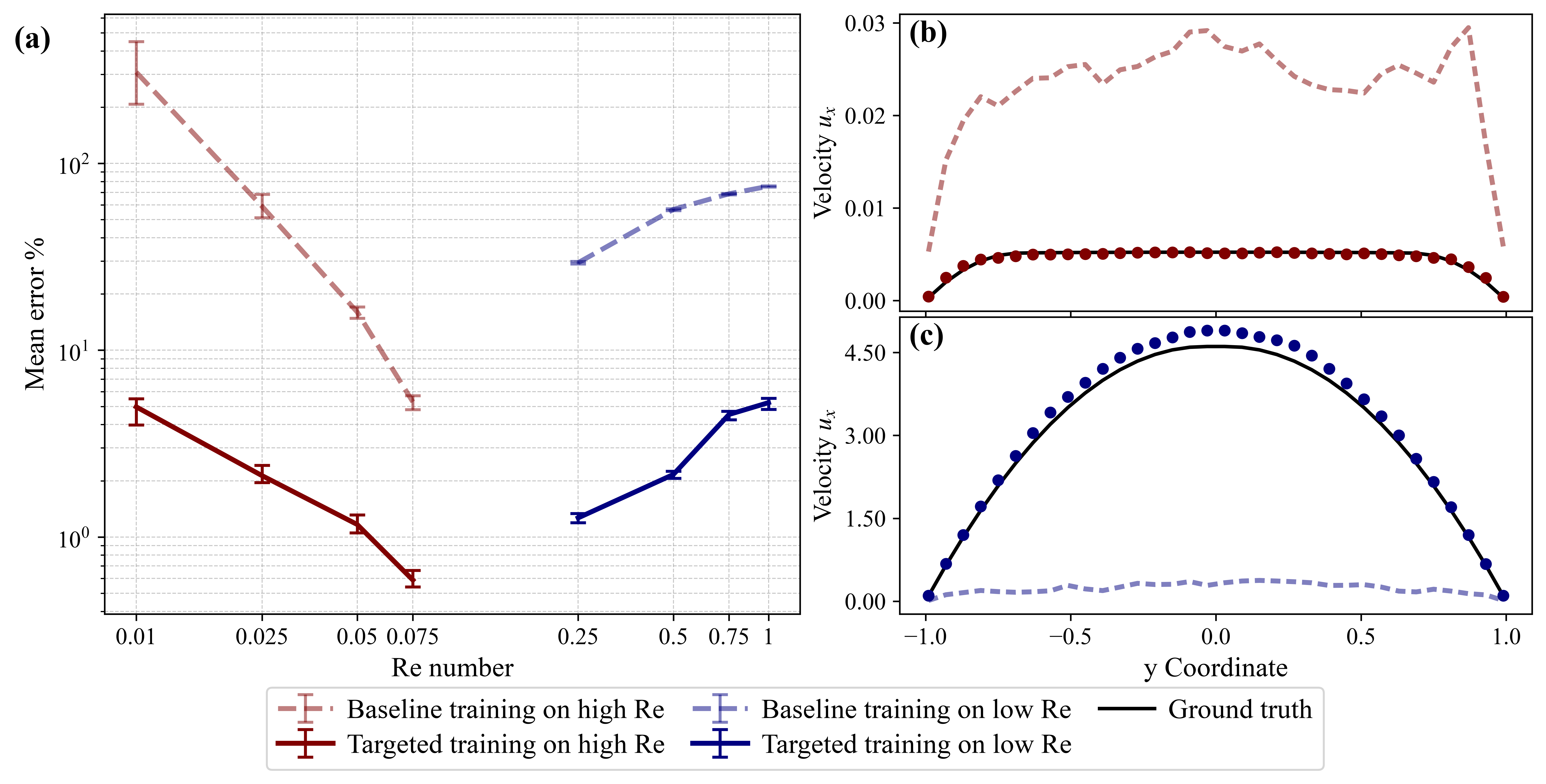}
    \caption{Effect of targeted sampling on extrapolation across the $Re$ range. (a) Mean prediction error versus $Re$ before (dashed) and after (solid) targeted sampling, for models trained on low $Re$ augmented with a few high-$Re$ corner cases (blue), and for models trained on high $Re$ augmented with a few low-$Re$ cases (red). (b) $Re = 0.01$, $Bi = 0.068$, model trained on high $Re$. (c) $Re = 1$, $Bi = 0.4$, model trained on low $Re$. Panels (b) and (c) compare the ground truth with predictions before and after targeted sampling, using the same line styles and colors as in (a).}
    \label{fig:Re_improved}
\end{figure}

The extrapolation failures identified in previous Sections reveal a consistent underlying limitation of RheOFormer: prediction accuracy degrades primarily when the training dataset does not expose the model to qualitatively distinct physical regimes. In all extrapolation scenarios considered, the breakdown does not stem from insufficient resolution within a given regime, but rather from the absence of representative examples that encode fundamentally different flow structures or velocity scales. This observation motivates a targeted sampling strategy in which a minimal number of carefully selected corner cases are added to the training set to span the missing dynamical behaviors.

\begin{figure}
    \centering
    \includegraphics[width=\linewidth]{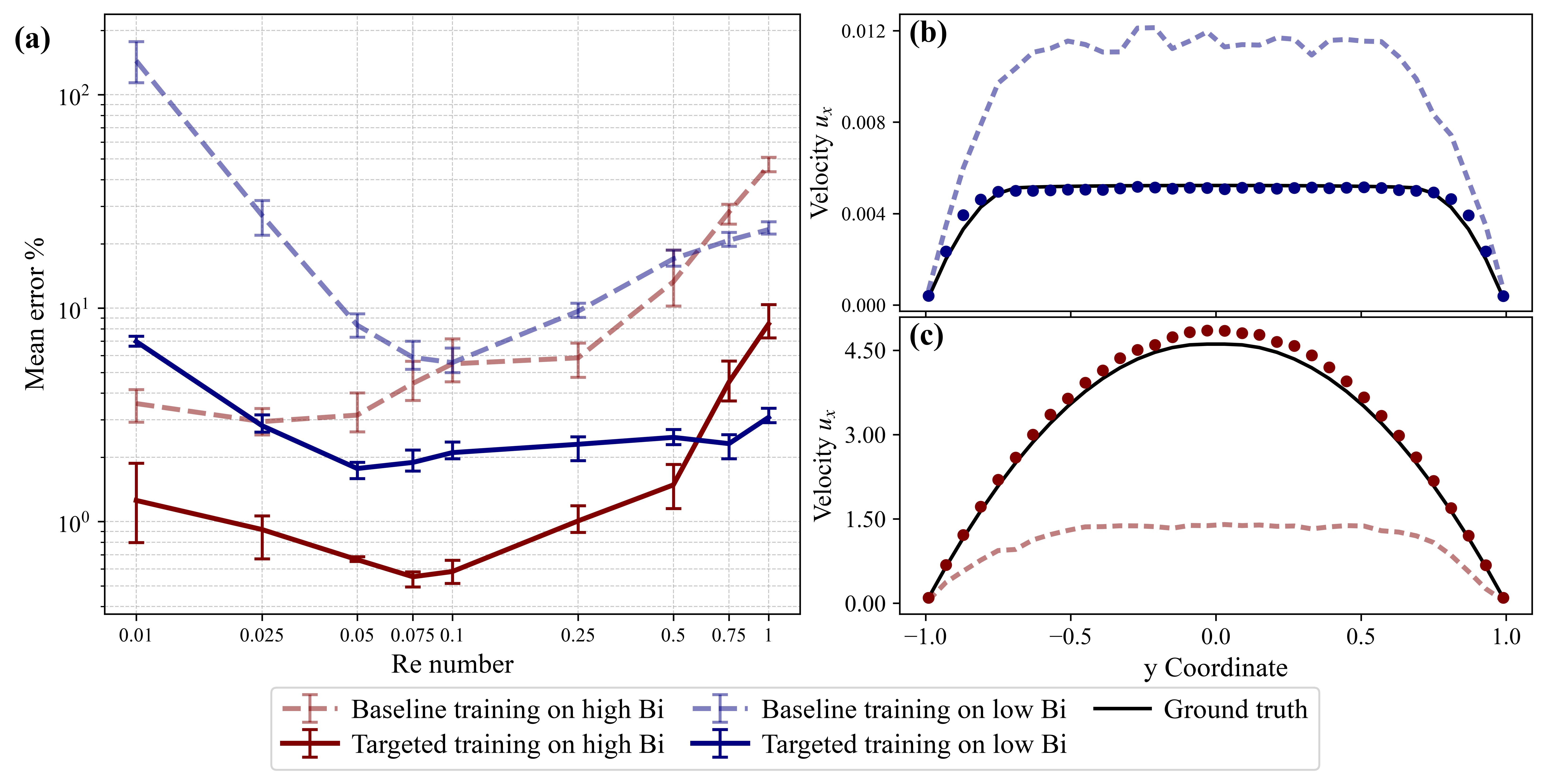}
    \caption{Effect of targeted sampling on extrapolation across the $Bi$ range. (a) Mean prediction error for each $Re$ before (dashed) and after (solid) adding a minimal number of corner cases at extreme $Bi$. (b) $Re = 0.01$, $Bi = 0.068$, model trained on low $Bi$ (blue). (c) $Re = 1.0$, $Bi = 0.4$, model trained on high $Bi$ (red). Panels (b) and (c) compare the ground truth with predictions before and after targeted sampling.}
    \label{fig:Bi_improved}
\end{figure}

We first revisit extrapolation across the $Re$ axis. When training is restricted to the low $Re$ regime, the model is exposed only to low magnitude velocity fields and consequently fails to predict higher $Re$ flows, as shown in Section \ref{Re_extrapolation}. To address this limitation, we augment the training set by adding a small number of representative high $Re$ samples. Specifically, for each previously unseen $Re=\{0.25,\,0.5,\,0.75,\,1.0\}$, we include only three cases corresponding to the minimum, midpoint, and maximum $Bi$ value. This results in a total of just 12 additional training samples. Despite this minimal increase in dataset size, the improvement is substantial.  As shown in Figure~\ref{fig:Re_improved}(a), the mean relative error decreases by more than an order of magnitude across the entire $Re$ range. The model is no longer limited by velocity scale mismatch and successfully recovers both the magnitude and shape of the steady state profiles, as illustrated in Figure~\ref{fig:Re_improved}(c). These results indicate that exposure to a small number of representative high $Re$ solutions is sufficient for the model to internalize the appropriate scaling behavior and transition between low  and high velocity regimes.

A complementary improvement is observed when training is initially restricted to the high $Re$ regime. In this case, the model tends to overestimate velocities and fails to reconstruct low $Re$ flows. By symmetry with the previous experiment, we augment the training set by adding only two samples per unseen $Re$ (corresponding to the minimum and maximum $Bi$ for $Re=\{0.01,\,0.025,\,0.05,\,0.075\}$), for a total of eight additional samples. As shown in Figures \ref{fig:Re_improved}(a) and (b), this minimal augmentation again leads to a dramatic reduction in extrapolation error.

An analogous pattern emerges for extrapolation across the $Bi$ axis. When training is limited to low $Bi$ flows, the model fails to predict strongly viscoplastic regimes characterized by wide plug regions and suppressed shear. In this case, the missing physics corresponds to the emergence of unyielded regions rather than changes in velocity magnitude. Remarkably, adding only a single training sample at the highest $Bi$ for each $Re$ (nine additional samples in total) is sufficient to restore accurate predictions across the entire $Bi$ range (Figure~\ref{fig:Bi_improved}(a), (b)). Conversely, when training is restricted to high $Bi$ regimes, the model struggles to recover the Newtonian-like behavior. Here, the absence of fully yielded reference states prevents the model from reconstructing high curvature velocity profiles. In this case, including a single Newtonian sample ($Bi=0$) for each $Re$ (again nine additional samples in total) leads to a dramatic improvement in accuracy (Figure~\ref{fig:Bi_improved}(a), (c)). This highlights the special role of the Newtonian limit as an extreme point in parameter space, enabling the operator to interpolate correctly between yielded and unyielded behaviors.

These results demonstrate that the dominant source of extrapolation error is not insufficient data volume or limited model capacity, but incomplete coverage of qualitatively distinct physical regimes. By adding a minimal number of strategically chosen corner cases that span the extremes of $Re$ and $Bi$, RheOFormer can generalize reliably across the full parameter space. While additional data naturally continue to improve performance, these findings show that even a very small number of representative samples can enable the model to internalize new physical behaviors and accurately interpolate between them.

\subsection{Effect of Input Temporal Information} \label{sec:temporal}

\begin{figure*}
\centering
\includegraphics[width=\linewidth]{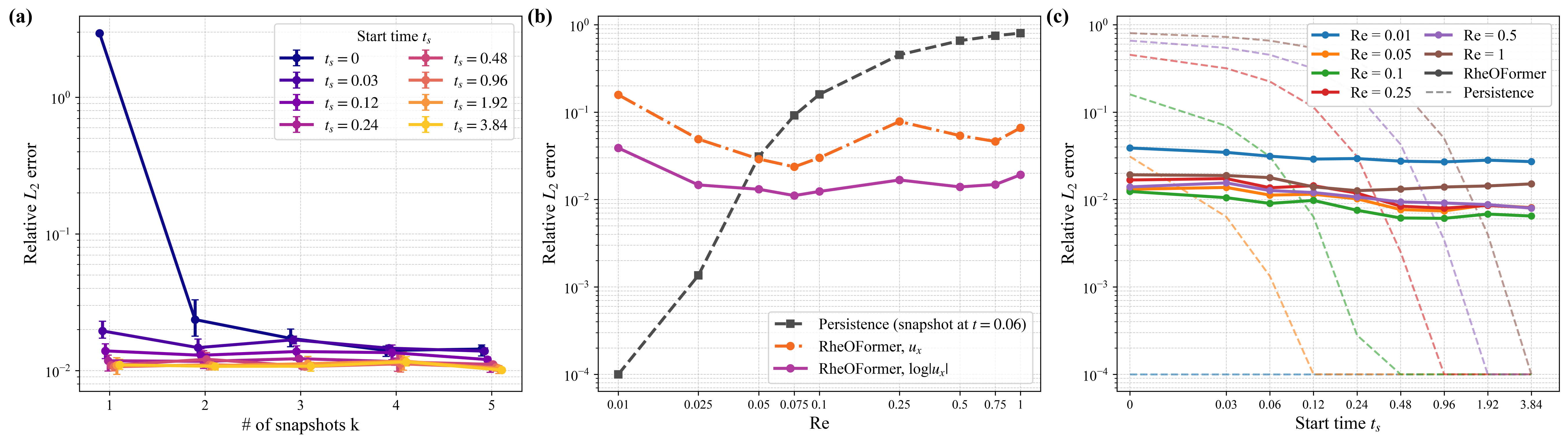}
\caption{Sensitivity of steady state prediction accuracy to the amount and timing of transient input. 
(a) Log-velocity model: error averaged over all $Re$ as a function of the number of input snapshots $k$, for input sequences starting at different times $t_s$. 
(b) Comparison with the persistence baseline, for $k = 3$ snapshots starting at $t_s = 0$: persistence (dashed), and RheOFormer trained on $u_x$ (dash-dotted) and on $\log|u_x|$ (solid). 
(c) Log-velocity model with a window of $k = 3$ snapshots starting at different times $t_s$: RheOFormer (solid) and persistence (dashed) for each $Re$.}
\label{fig:temporal_input}
\end{figure*}

A natural question in operator learning for dynamical systems is whether providing additional temporal information improves the model’s ability to infer long time or steady state behavior. Intuitively, access to multiple early-time snapshots may help the model better infer the governing dynamics and thereby enhance steady state predictions. In this subsection, we explicitly test this hypothesis by systematically varying both the amount and the timing of the transient information supplied to RheOFormer while keeping all other aspects of the training procedure fixed. 
Specifically, the model receives $k$ consecutive snapshots of the transient flow evolution, uniformly spaced in time with $\Delta t = 0.03$ and starting at time $t_s$, so that the last input snapshot is taken at $t_{in} = t_s + (k-1)\,\Delta t$, and is asked to predict the steady state velocity field. The number of input frames is varied as $k = 1, 2, 3, 4,$ and $5$, and the start time is varied between $t_s = 0$ and $t_s = 3.84$, so that the inputs range from a barely developed flow to an essentially steady one. All runs use the log-velocity configuration of Section~\ref{Improving} and the training set of five cases per $Re$ (Sections~\ref{plateau_error}--\ref{Improving}); errors are reported as the relative $L_2$ error over the held-out cases, averaged over three training seeds. The model architecture, loss function, and training protocol are otherwise unchanged.

Figure~\ref{fig:temporal_input}(a) summarizes the error, averaged over all $Re$, as a function of the number of input snapshots $k$. For input sequences starting at $t_s = 0$, a pronounced reduction in error is observed when increasing the input from a single snapshot to $k=3$, by more than two orders of magnitude. Beyond this point, the error plateaus at $0.014$--$0.017$, and the inclusion of additional snapshots yields no systematic improvement; the remaining variations are comparable to the variation between training seeds.
As expected, the model does not require access to the initial condition itself.
All simulations in the dataset start from a quiescent state with zero velocity, so that the initial snapshot carries no case-specific information. 
This explains the failure of $k=1$ at $t_s = 0$, where the input is identical for all cases. 
For every start time $t_s \geq 0.03$, a single snapshot already yields a good prediction for the steady state ($2 \%$ error at $t_s = 0.03$ and $1\%$ for $t_s \geq 0.12$), and increasing the input to five snapshots lowers the error by at most about $30\%$. This saturation indicates that RheOFormer does not rely on an extended temporal history: as soon as the input contains a partially developed flow field, one or a few snapshots are sufficient to identify the flow regime and infer its steady state. Since a single snapshot performs nearly as well as several, the relevant information is carried by the spatial structure of the developing field, namely its magnitude and the extent of the plug region, rather than by its temporal evolution.

To assess whether the model genuinely infers the long time behavior rather than reproducing its input, we compare it with the simplest possible predictor, which we refer to as persistence: the last input snapshot is taken directly as the prediction of the steady state, $u_x^{\infty} \approx u_x(x,y,t_{in})$. 
The error of this baseline is a direct measure of how far the flow still is from its steady state at $t_{in}$. Figure~\ref{fig:temporal_input}(b) shows this comparison for $k = 3$ snapshots starting at $t_s = 0$, i.e. for inputs covering only $t = 0$--$0.06$.
Because the viscous term scales as $1/Re$, the time required for the flow to develop increases with $Re$, and the persistence error rises accordingly, from below $10^{-4}$ at $Re = 0.01$, where the flow is very close to steady at $t = 0.06$, to $0.80$ at $Re = 1$, where it has barely started. The log-velocity model, in contrast, maintains an error of $0.011$--$0.019$ over the entire range, independently of how far the flow has developed. For $Re \geq 0.05$ it outperforms persistence, and the advantage grows with $Re$, reaching a factor of about $40$--$50$ for $Re \geq 0.5$. In this regime the operator performs genuine forecasting: from inputs that are far from the steady state, it infers the long time behavior rather than copying the input. 
The model trained on the raw velocity follows the same trend, only with larger errors.
At $Re \leq 0.025$ the inputs are very close to the steady state and the task reduces to reconstructing a steady field from steady inputs, so persistence is essentially exact, while the model error ($0.015$ and $0.039$) reflects the residual error.

Finally, Figure~\ref{fig:temporal_input}(c) extends the comparison with persistence to all input windows, by shifting the $k = 3$ window later in time. The persistence error depends strongly on the start time and decreases rapidly once the flow approaches its steady state, which occurs later for larger $Re$, whereas the RheOFormer error remains nearly constant.
As a result, RheOFormer outperforms persistence as long as the last input snapshot is taken before $t_{in} \approx Re$, which corresponds to the time required for the flow to develop. The advantage observed in Figure~\ref{fig:temporal_input}(b) is therefore not specific to the chosen window, but holds throughout the development of the flow. Moreover, the insensitivity of the model to the timing of the input holds for each $Re$ individually and not only on average: the error remains around $0.5\%$--$2\%$ for $Re \geq 0.05$ and $2\%$--$4\%$ at $Re = 0.01$ for all start times.
Hence, the model does not need the snapshots to be taken at a particular time and it can predict the steady state with the same accuracy from a window at the very beginning of the transient as from a window close to the steady state.
The practical value of this ability is greatest precisely in the early-time regime, where only a short initial portion of the transient is available.

For the steady channel flows considered here, these results are consistent with the conclusions of the preceding sections: the composition of the training dataset in parameter space and the representation of distinct physical regimes are far more critical than the amount of temporal information provided. A single informative snapshot, or a few snapshots taken early in the transient, well before the flow has developed, are sufficient for an accurate steady state prediction.

\section{Conclusion}\label{sec13}
In this work, we conducted a systematic examination of RheOFormer, an attention-based neural operator, for predicting steady state viscoplastic channel flows governed by Bingham physics. Beyond evaluating baseline accuracy, our objective was to understand the fundamental mechanisms that enable or hinder generalization across the full $Re$–$Bi$ parameter space, and to establish a practical and data efficient methodology for enabling accurate predictions using minimal data. Rather than focusing solely on performance metrics, this study emphasizes understanding why and where operator learning succeeds or fails for viscoplastic systems. The findings of this study reveal clear numerical characteristics of viscoplastic flows that shape model performance and motivate improvements in both input representation and loss design. Through this analysis, several methodological insights emerged that together form a generalizable roadmap for training RheOFormer on different physical systems.

Bingham flows introduce extended regions of nearly zero velocity gradients associated with unyielded plug regions. 
A central observation is that these low shear zones are numerically subtle and weakly penalized by standard loss functions, causing standard representations to underemphasize the plateau region and leading to systematic errors in the predicted plateau height when training data are limited. Importantly, this failure mode is not attributable to insufficient model capacity, but rather to the inherent numerical structure of viscoplastic flows, where meaningful spatial variations are concentrated in thin boundary layers. Recognizing this led to a series of methodological contributions aimed at improving plateau region accuracy. By augmenting the input representation with gradient-based features and, in particular, logarithmic transformations, we demonstrated that the numerical salience of low gradient regions can be significantly enhanced.  
The log-velocity representation consistently amplified weak shear information, enabling the model to recover the correct plateau velocity even in strongly viscoplastic regimes, and gave the lowest error of all strategies. 
The raw velocity gradient as an additional input improved the plateau prediction at high $Re$ but not at low $Re$.
In parallel, we evaluated several loss formulations and found that Sobolev loss provides a stable and physically grounded mechanism for jointly penalizing velocity and gradient errors, with the clearest gain in the highest-$Bi$ case at $Re=1$. 
At $Re=0.01$, however, where both the velocity magnitudes and their gradients are extremely small, none of the modifications fully resolves the plateau error, and accurate prediction of plug-dominated flows in this regime remains challenging.
These representations and loss strategies offer a principled approach for addressing low gradient failure modes in neural operators applied to non-Newtonian flows.

In addition to improving accuracy within the training domain, we systematically probed the model’s ability to extrapolate along both the $Re$ and $Bi$ dimensions. Our results revealed two important aspects of extrapolation behavior in RheOFormer. First, models trained only on low $Re$ flows were unable to reconstruct the higher magnitude velocity profiles, while models trained exclusively on high $Re$ flows systematically overestimated velocities and failed to recover the lower velocity magnitudes. 
This underscores the importance of the numerical range of values encountered during training: downward extrapolation performs slightly better because high-$Re$ training sets already contain moderately small velocities at high $Bi$, whereas upward extrapolation requires predicting velocity magnitudes and gradients outside the distribution observed during training. 
Second, our study highlights the necessity of covering the full spectrum of physical behaviors in the $Bi$ dimension. 
Models trained only on small $Bi$ cannot reconstruct viscoplastic flows with pronounced plug regions, while models trained solely on large $Bi$ struggle to reproduce Newtonian-like curvature in fully yielded states.
Although fully yielded profiles lack a yield surface and are structurally simpler, their velocity magnitudes and shear rates at high $Re$ exceed those encountered in the plug-dominated training data, making extrapolation toward them difficult.
These findings emphasize a central practical implication: RheOFormer does not extrapolate reliably across dynamically distinct regimes unless the underlying physical and numerical space is adequately sampled. Motivated by this, we identified targeted sampling strategies that substantially mitigate these failures by selecting a minimal yet representative set of training cases that spans the extreme behaviors of the system.

Finally, the temporal input study showed that the model does not rely on an extended temporal history: apart from the uninformative initial condition, a single snapshot is nearly as informative as several, and the accuracy is nearly independent of when the input window is taken. Compared with the trivial predictor that copies the last input, RheOFormer is up to about 50 times more accurate while the flow is still developing, i.e. for input times up to $t \approx Re$.

The channel flow studied here is deliberately simple with flat and fixed yield surfaces, unidirectional velocity, and the steady profile that depends only on Bi/Re. Many viscoplastic flows of practical interest are much more complicated with dynamic heterogenous stress fields or deformable interfaces, where yield surfaces interact with a free surface and more complex boundary conditions. In the latter, the domain itself evolves in time, and the material stops in a shape set by the balance of driving forces and the yield stress\cite{francca2024elasto,van2023viscoplastic}. In these problems, fundamental to applications such as 3D printing, the quantity of interest is often the interface shape and the arrest time rather than a velocity field. Such flows are natural next tests for the strategies identified here, but would require alternative strategies for instance to the logarithmic velocity transform.

Overall, this study provides both conceptual understanding and actionable guidance for training RheOFormer on complex rheological systems with minimal data. 
The improvements introduced (log-scaled representations, gradient aware losses, and targeted sampling) are not specific to Bingham flow and are expected to transfer to other classes of complex fluids or general nonlinear PDEs where low gradient structures or sharp transitions challenge standard learning approaches. By clarifying the interplay between data structure, numerical representation, and operator learning performance, this work establishes a foundation for reliably training RheOFormer for a broad spectrum of rheological and physics-based modeling applications.

\appendix

\renewcommand{\theequation}{\thesection.\arabic{equation}}
\setcounter{section}{0}
\refstepcounter{section}
\setcounter{equation}{0}

\section*{Appendix \thesection. Numerical setup and validation}
\label{section:app:numerical_details_validation}
To generate the training dataset, two-dimensional pressure-driven Bingham flows were simulated using the Basilisk flow solver. The governing equations were nondimensionalized using the channel half-width $H$ as the characteristic length scale and a reference velocity $U$ as the velocity scale, giving
\begin{equation}
    \mathbf{x}=H\bar{\mathbf{x}},
    \qquad
    t=\frac{H}{U}\bar{t},
    \qquad
    \mathbf{u}=U\bar{\mathbf{u}},
    \qquad
    p=\rho U^2\bar{p},
\label{eq:rescalings}
\end{equation}
where $\rho$, $\mathbf{x}$, $t$, $\mathbf{u}$ and $p$ are, respectively: density, position, time, velocity and pressure. Values with a bar on top are nondimensional.

The resulting dimensionless equations are
\begin{equation}
    \frac{D\bar{\mathbf{u}}}{D\bar{t}}
    =
    -\bar{\nabla}\bar{p}
    +
    \bar{\nabla}\cdot\bar{\boldsymbol{\tau}},
    \qquad
    \bar{\nabla}\cdot\bar{\mathbf{u}}=0,
\end{equation}
with
\begin{equation}
    \bar{\boldsymbol{\tau}}
    =
    2\left[
    \frac{1}{Re}
    +
    \frac{Bi}{Re}
    \frac{1}{2\|\bar{\mathbf{D}}\|}
    \right]
    \bar{ \mathbf{D} },
\label{eq:stress_bingham}
\end{equation}
where
\begin{equation}
    \mathbf{D}
    =
    \frac{1}{2}
    \left(
    \bar{\nabla}\bar{\mathbf{u}}
    +
    \bar{\nabla}\bar{\mathbf{u}}^T
    \right)
\end{equation}
is the rate-of-deformation tensor, and the tensor norm is defined as
\begin{equation}
    \|\bar{ \mathbf{D} }\|
    =
    \left(
    \frac{1}{2}\bar{ \mathbf{D} }:\bar{ \mathbf{D} }
    \right)^{1/2}.
\end{equation}

The Reynolds and Bingham numbers are defined as
\begin{equation}
    Re=\frac{\rho U H}{\mu},
    \qquad
    Bi=\frac{\tau_y H}{\mu U},
\end{equation}
where $\rho$, $\mu$, and $\tau_y$ denote the density, plastic viscosity, and yield stress, respectively.

The flow is driven by a constant streamwise pressure gradient. We define the dimensionless pressure-gradient parameter $\bar{K}$ according to
\begin{equation}
    k=\frac{\rho U^2}{H}\bar{K},
\end{equation}
where $k=-\partial p/\partial x$. Thus, $\bar{K}=-\partial \bar{p}/\partial \bar{x}$. 

With these nondimensionalization choices, the flow admits the following analytical solution for the velocity profile
\begin{equation}
u_x(y) = 
     \begin{cases}
       \frac{1}{2}Re\, \bar{K}(1-y^2) - Bi(y + 1), & \quad\text{if}\quad y \leq -\frac{Bi}{Re}\frac{1}{\bar{K}},\\
       %%%%%%%%%%%%%%%%
       \frac{1}{2}Re\, \bar{K} - Bi + \frac{1}{2}\frac{Bi^2}{Re}\frac{1}{\bar{K}}, &\quad\text{if}\quad |y| < \frac{Bi}{Re}\frac{1}{\bar{K}},\\
       %%%%%%%%%%%%%%%%
       \frac{1}{2}Re\, \bar{K}(1-y^2) + Bi(y - 1), &\quad\text{if}\quad y \geq \frac{Bi}{Re}\frac{1}{\bar{K}}.\\
       %%%%%%%%%%%%%%%%
     \end{cases}
\label{eq:analytical_solution}
\end{equation}

The computational domain is the dimensionless rectangular channel $(\bar{x},\bar{y})\in[0,8]\times[-1,1]$, where $\bar{y}=\pm1$ correspond to the channel walls. No-slip boundary conditions are imposed at the upper and lower walls. At the inlet and outlet, the pressure is prescribed and the normal velocity gradient is set to zero, corresponding to the pressure-driven boundary conditions implemented in Basilisk.

The ideal Bingham constitutive relation is regularized using the Papanastasiou model \cite{papanastasiou1987flows}. Specifically, the apparent viscosity in Eq.~\eqref{eq:stress_bingham} is replaced by
\begin{equation}
\label{analytical}
    \mu_{\mathrm{app}}
    =
    \frac{1}{Re}
    +
    \frac{Bi}{Re}
    \frac{\left[ 1-\exp(-\|\bar{\mathbf{D}}\|/\epsilon) \right] } 
    {2\|\bar{\mathbf{D}}\|},
\end{equation}
where the regularization parameter is kept fixed at $\epsilon=10^{-4}$.

The simulations are performed on a uniform Cartesian grid with spacing $\Delta\bar{x}=\Delta\bar{y}  = 8/2^8 = 0.03125$. Each simulation is advanced until a steady state is reached, defined by
\begin{equation}
    \max_{\mathbf{x}}
    \left|
    \bar{u}_x^{n+1}(\mathbf{x})
    -
    \bar{u}_x^{n}(\mathbf{x})
    \right|
    <10^{-8}.
\end{equation}

The streamwise velocity field $\bar{u}_x(\bar{x},\bar{y},\bar{t})$ is recorded throughout each simulation. For use in the machine-learning model, the resulting velocity fields are downsampled to a $100\times100$ spatial grid and subsequently to the $34\times34$ grid used by the model.

\begin{figure}
\centering
\includegraphics[width=\linewidth]{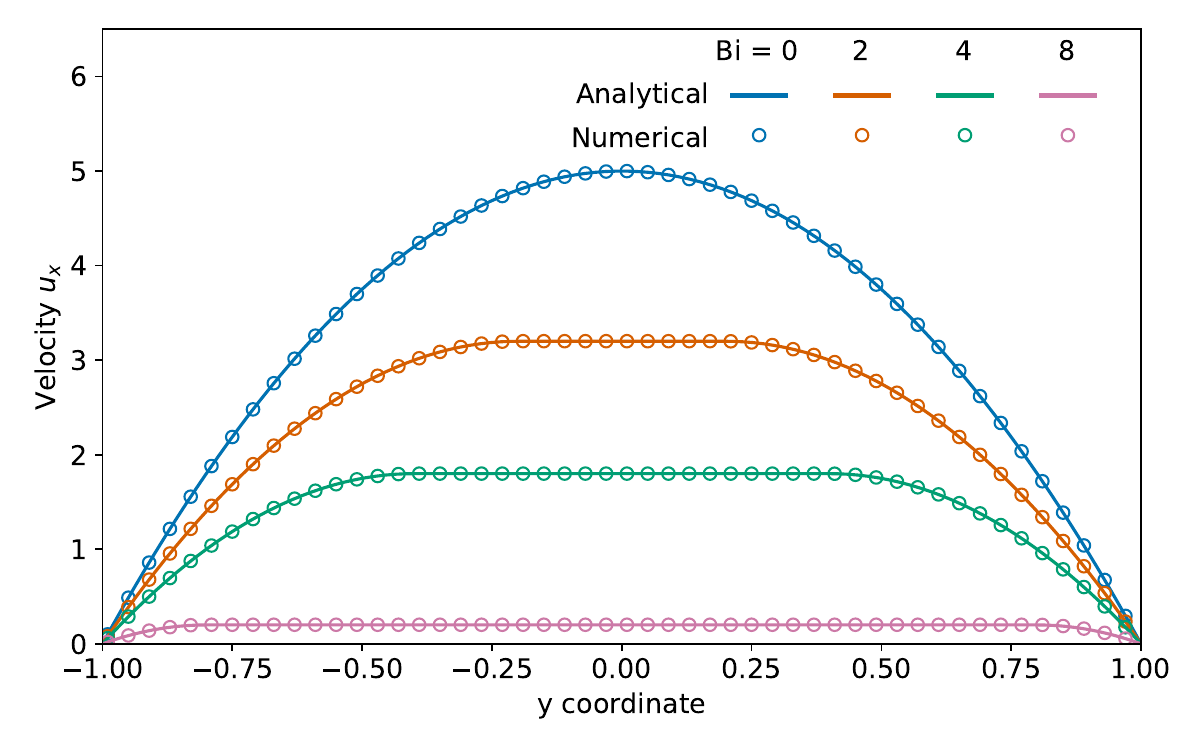}
\caption{Validation of the numerical data obtained from Basilisk against analytical solutions. All simulations shown use $Re=1$, $\bar{K}=10$, and $\epsilon=10^{-4}$. Four values of $Bi$ are shown, illustrating that the Basilisk solution captures the yielding well, even for highly plugged flows.}
\label{fig:app:validation}
\end{figure}

In order to demonstrate that the data used in training correctly captures the expected yielding behavior, we show in Fig.~\ref{fig:app:validation} a comparison between analytical solutions from Eq.~\eqref{eq:analytical_solution} and numerical solutions obtained with Basilisk. We observe good agreement between these solutions, which validates that the data used in our training dataset contains a good representation of the yielding behavior of Bingham fluids.

% Appendix sections are coded under \verb+\appendix+.

% \verb+\printcredits+ command is used after appendix sections to list 
% author credit taxonomy contribution roles tagged using \verb+\credit+ 
% in frontmatter.

\subsubsection*{Declaration of competing interest}
The authors declare that there is no conflict of interest.

\subsubsection*{Acknowledgements}

M.S. and S.J. would like to acknowledge the support of DoD MURI Award N00014-23-1-2499.

\printcredits

% \bibliographystyle{cas-model2-names}
\bibliographystyle{elsarticle-num-names}

\bibliography{cas-refs}

\end{document}